\documentclass{article}
\usepackage{arxiv}
\pdfoutput=1 
\usepackage[utf8]{inputenc} % allow utf-8 input 
\usepackage[T1]{fontenc}    % use 8-bit T1 fonts 
\usepackage{hyperref}       % hyperlinks 
\usepackage{url}            % simple URL typesetting
\usepackage{booktabs}       % professional-quality tables
\usepackage{amsfonts}       % blackboard math symbols
\usepackage{nicefrac}       % compact symbols for 1/2, etc.
\usepackage{microtype}      % microtypography
\usepackage{lipsum}
  
\usepackage{times}  
\usepackage{latexsym} 
\usepackage{graphics,xspace,epsfig}  
\usepackage{amsmath}
\usepackage{subfigure}
\usepackage{amsfonts}  
\usepackage{amssymb}  
\usepackage{enumerate}
\usepackage{varwidth} 
\usepackage{multirow} 
\usepackage{url}  
\usepackage[flushleft]{threeparttable} 
\usepackage{flushend}   
\usepackage{xcolor,colortbl} 
\usepackage{adjustbox} 
\usepackage{enumitem}
  
\title{A Question-Entailment Approach \\to Question Answering}  

\author{
 Asma {Ben Abacha}~~~~~~~~~~~~~~~~~~~~~~~~~ Dina Demner-Fushman\\ 
  \texttt{\small{benabachaa@nih.gov ~~~~~~~~ ddemner@mail.nih.gov}}    \\\\
    Lister Hill Center, U.S. National Library of Medicine, \\ U.S. National Institutes of Health, Bethesda, MD 
}  
  
\begin{document}
\maketitle

\begin{abstract}
One of the challenges in large-scale information retrieval (IR) is to develop fine-grained and domain-specific methods to answer natural language questions. Despite the availability of numerous sources and datasets for answer retrieval, Question Answering (QA) remains a challenging problem due to the difficulty of the question understanding and answer extraction tasks. One of the promising tracks investigated in QA is to map new questions to formerly answered questions that are ``similar''. In this paper, we propose a novel QA approach based on Recognizing Question Entailment (RQE) and we describe the QA system and resources that we built and evaluated on real medical questions. First, we compare machine learning and deep learning methods for RQE using different kinds of datasets, including textual inference, question similarity and entailment in both the open and clinical domains. Second, we combine IR models with the best RQE method to select entailed questions and rank the retrieved answers. To study the end-to-end QA approach, we built the MedQuAD collection of 47,457 question-answer pairs from trusted medical sources, that we introduce and share in the scope of this paper. Following the evaluation process used in TREC 2017 LiveQA, we find that our approach exceeds the best results of the medical task with a 29.8\% increase over the best official score. The evaluation results also support the relevance of question entailment for QA and highlight the effectiveness of combining IR and RQE for future QA efforts. Our findings also show that relying on a restricted set of reliable answer sources can bring a substantial improvement in medical QA.  
\end{abstract}  
 
 
%====================== 
\section{Introduction}  
\label{sec:into}  
%======================

With the availability of rich data on users' locations, profiles and search history, personalization has become the leading trend in large-scale information retrieval. However, efficiency through personalization is not yet the most suitable model when tackling domain-specific searches. This is due to several factors, such as the lexical and semantic challenges of domain-specific data that often include advanced argumentation and complex contextual information, the higher sparseness of relevant information sources, and the more pronounced lack of similarities between users' searches. 

A recent study on expert search strategies among healthcare information professionals \cite{study2017} showed that, for a given search task, they spend an average of 60 minutes per collection or database, 3 minutes to examine the relevance of each document, and 4 hours of total search time. When written in steps, their search strategy spans over 15 lines and can reach up to 105 lines.   
  
With the abundance of information sources in the medical domain, consumers are more and more faced with a similar challenge, one that needs dedicated solutions that can adapt to the heterogeneity and specifics of health-related information.

Dedicated Question Answering (QA) systems are one of the viable solutions to this problem as they are designed to understand natural language questions without relying on external information on the users.  

In the context of QA, the goal of Recognizing Question Entailment (RQE) is to retrieve answers to a \textit{premise question} ($PQ$) by retrieving inferred or entailed questions, called \textit{hypothesis questions} ($HQ$) that already have associated answers. Therefore, we define the entailment relation between two questions as: a question $A$ \textbf{entails} a question $B$ if every answer to $B$ is also a \textbf{correct answer} to $A$ \cite{BenAbacha_AMIA_2016}.  

RQE is particularly relevant due to the increasing numbers of similar questions posted online \cite{SimQ:J2015} and its ability to solve differently the challenging issues of question understanding and answer extraction. In addition to being used to find relevant answers, these resources can also be used in training models able to recognize inference relations and similarity between questions.  

Question similarity has recently attracted international challenges \cite{semEval_2016_ref,semEval_2017_ref} and several research efforts proposing a wide range of approaches, including Logistic Regression, Recurrent Neural Networks (RNNs), Long Short Term Memory cells (LSTMs), and Convolutional Neural Networks (CNNs)   \cite{Santos:ACL2015,Romeo:Coling2016,BenAbacha_AMIA_2016,Lei:NAACL2016}.   
   
In this paper, we study question entailment in the medical domain and the effectiveness of the end-to-end RQE-based QA approach by evaluating the relevance of the retrieved answers. Although entailment was attempted in QA before \cite{Harabagiu:2006,QA-Entail-RANLP:2009,Celikyilmaz:ACL09}, as far as we know, we are the first to introduce and evaluate a full medical question answering approach based on question entailment for free-text questions. Our contributions are:  
\begin{enumerate} 
\item A study of machine learning and deep learning approaches to RQE using different kinds of datasets, including textual inference, question similarity and entailment in both the open and clinical domains.
\item A collection of 47,457 medical question-answer pairs with additional annotations, constructed from trusted sources such as NIH websites. We make this resource publicly available\footnote{\url{https://github.com/abachaa/MedQuAD}}.  
\item A new QA approach based on question entailment. Our approach uses IR models to retrieve question candidates and the RQE model to identify entailed questions and return their answers.  
\item An evaluation of the RQE-based QA system on TREC 2017 LiveQA medical questions \cite{liveQA-Med-overview-2017}. Results showed that our approach exceeds the best official score on the medical task using only the collection of 47K QA pairs as answers source.   
\end{enumerate}     

The next section is dedicated to related work on question answering, question similarity and entailment. In Section~\ref{sec3:RQE}, we present two machine learning (ML) and deep learning (DL) methods for RQE and compare their performance using open-domain and clinical datasets. Section~\ref{sec4:collection}  describes the new collection of medical question-answer pairs. In Section~\ref{sect:system}, we describe our RQE-based approach for QA. Section~\ref{sec6:eval} presents our evaluation of the retrieved answers and the results obtained on TREC 2017 LiveQA medical questions.                       
 %======================            
\section{Background}   
\label{sec:back}     
%======================    

In this section we define the RQE task and describe related work at the intersection of question answering, question similarity and textual inference.  

\subsection{Task Definition}          
\label{sec:taskDef}

The definition of Recognizing Question Entailment (RQE) can have a significant impact on QA results. In related work, the meaning associated with Natural Language Inference (NLI) varies among different tasks and events. For instance, Recognizing Textual Entailment (RTE) was addressed by the PASCAL challenge \cite{Book:Dagan:2013}, where the entailment relation has been assessed manually by human judges who selected relevant sentences "entailing" a set of hypotheses from a list of documents returned by different Information Retrieval (IR) methods. In another definition, the  Stanford  Natural  Language  Inference corpus SNLI \cite{snli:emnlp2015}, used three classification labels for the relations between two sentences: entailment, neutral and contradiction. For the entailment label, the annotators who built the corpus were presented with an image and asked to write a caption ``that is a definitely true description of the photo''. For the neutral label, they were asked to provide a caption ``that might be a true description of the label''. They were asked for a caption that ``is definitely a false description of the photo'' for the contradiction label.    

More recently, the multiNLI corpus \cite{multi-nli:2017} was shared in the scope of the RepEval 2017 shared task\footnote{\url{https://repeval2017.github.io/shared}} \cite{RepEval:2017}. To build the corpus, annotators were presented with a premise text and asked to write three sentences. One novel sentence, which is ``necessarily true or appropriate in the same situations as the premise,'' for the entailment label, a sentence, which is ``necessarily false or inappropriate whenever the premise is true,'' for the contradiction label, and a last sentence, ``where neither condition applies,'' for the neutral label. 
 
Whereas these NLI definitions might be suitable for the broad topic of text understanding, their relation to practical information retrieval or question answering systems is not straightforward. 

In contrast, RQE has to be tailored to the question answering task. For instance, if the premise question is "looking for cold medications for a 30 yo woman", a RQE approach should be able to consider the more general (less restricted) question "looking for cold medications" as relevant, since its answers are relevant for the initial question, whereas "looking for medications for a 30 yo woman" is a useless contextualization. The entailment relation we are seeking in the QA context should include relevant and meaningful relaxations of contextual and semantic constraints (cf. Section~\ref{sec:RQEDef}).

\subsection{Related Work on Question Answering}    
     
Classical QA systems face two main challenges related to question analysis and answer extraction. Several QA approaches were proposed in the literature for the open domain \cite{CFO_QA_ACL2016,QASurvey:17} and the medical domain \cite{Athenikos:BioQASurvey2010,MEANS2015,liveQA-CMU-LiveMedQA}.  A variety of methods were developed for question analysis, focus (topic) recognition and question type identification \cite{Watson:QA,BenAbacha:IHI2012,MrabetKRD16,MOMTAZI2018380}. Similarly, many different approaches tackled document or passage retrieval and answer selection and (re)ranking \cite{WangQA2007,Surdeanu:CLJ2011,TymoshenkoAndMoschitti:CIKM2015}.  
        
An alternative approach consists in finding similar questions or FAQs that are already answered \cite{jijkoun2005retrieving,wang2009syntactic}. One of the earliest question answering systems based on finding similar questions and re-using the existing answers was FAQ FINDER \cite{burke1997question}. Another system that complements the existing Q\&A services of NetWellness\footnote{\url{http://netwellness.org/}} is SimQ \cite{SimQ:J2015}, which allows retrieval of similar web-based consumer health questions. SimQ uses syntactic and semantic features to compute similarity between questions, and UMLS \cite{UMLS_Ref} as a standardized semantic knowledge source. The system achieves 72.2\% precision, 78.0\% recall and 75.0\% F-score on NetWellness questions. However, the method was evaluated only on one question similarity dataset, and the retrieved answers were not evaluated. 
 
The aim of the medical task at TREC 2017 LiveQA was to develop techniques for answering complex questions such as consumer health questions, as well as to identify relevant answer sources that can comply with the sensitivity of medical information retrieval.  
  
The CMU-OAQA system \cite{liveQA-CMU-OAQA} achieved the best performance of 0.637 average score on the medical task by using an attentional encoder-decoder model for paraphrase identification and answer ranking. The Quora question-similarity dataset was used for training. The PRNA system \cite{liveQA-PRNA} achieved the second best performance in the medical task with 0.49 average score using Wikipedia as the first answer source and Yahoo and Google searches as secondary answer sources. Each medical question was decomposed into several subquestions. To extract the answer from the selected text passage, a bi-directional attention model trained on the SQUAD dataset was used. 
 
Deep neural network models have been pushing the limits of performance achieved in QA related tasks using large training datasets. The results obtained by CMU-OAQA and PRNA showed that large open-domain datasets   were beneficial for the medical domain. However, the best system (CMU-OAQA) relying on the same training data obtained a score of 1.139 on the LiveQA open-domain task.     

While this gap in performance can be explained in part by the discrepancies between the medical test questions and the open-domain questions, it also highlights the need for larger medical datasets to support deep learning approaches in dealing with the linguistic complexity of consumer health questions and the challenge of finding correct and complete answers.              
 
Another technique was used by ECNU-ICA team \cite{liveQA-ECNU} based on learning question similarity via two long short-term memory (LSTM) networks applied to obtain the semantic representations of the questions. To construct a collection of similar question pairs, they searched community question answering sites such as Yahoo! and Answers.com.  In contrast, the ECNU-ICA system achieved the best performance of 1.895 in the open-domain task but an average score of only 0.402 in the medical task. As the ECNU-ICA approach also relied on a neural network for question matching, this result shows that training attention-based decoder-encoder networks on the Quora dataset generalized better to the medical domain than training LSTMs on similar questions from Yahoo! and Answers.com. 
 
The CMU-LiveMedQA team \cite{liveQA-CMU-LiveMedQA} designed a specific system for the medical task. Using only the provided training datasets and the assumption that each question contains only one focus, the CMU-LiveMedQA system obtained an average score of 0.353. They used a convolutional neural network (CNN) model to classify a question into a restricted set of 10 question types and crawled "relevant" online web pages to find the answers. However, the results were lower than those achieved by the systems relying on finding similar answered questions. These results support the relevance of similar question matching for the end-to-end QA task as a new way of approaching QA instead of the classical QA approaches based on Question Analysis and Answer Retrieval.

%=======================================================
\subsection{Related Work on Question Similarity and Entailment}    
%======================================================= 

Several efforts focused on recognizing similar questions. Jeon et al. \cite{Jeon:2005} showed that a retrieval model based on translation probabilities learned from a question and answer archive can recognize semantically similar questions. Duan et al. \cite{duan2008searching} proposed a dedicated language modeling approach for question search, using question \textit{topic} (user's interest) and question \textit{focus} (certain aspect of the topic).         
  
Lately, these efforts were supported by a task on Question-Question similarity introduced in the community QA challenge at SemEval (task 3B) \cite{semEval_2016_ref}. Given a new question, the task focused on reranking all  similar  questions  retrieved  by  a  search  engine, assuming that the answers to the similar questions will be correct answers for the new question. Different machine learning and deep learning approaches were tested in the scope of SemEval 2016 \cite{semEval_2016_ref} and 2017 \cite{semEval_2017_ref} task 3B. The best performing system in 2017 achieved a MAP of 47.22\% using supervised Logistic Regression that combined different unsupervised similarity measures such as Cosine and Soft-Cosine \cite{SimBow:SemEvalBest2017}. The second best system achieved 46.93\% MAP with a learning-to-rank method using Logistic Regression and a rich set of features including lexical and semantic features as well as embeddings generated by different neural networks (siamese, Bi-LSTM, GRU and CNNs) \cite{SemEval2017-SecondT}. In the scope of this challenge, a dataset was collected from Qatar Living forum for training. We refer to this dataset as \textit{SemEval-cQA}\footnote{\url{http://alt.qcri.org/semeval2017/task3/index.php?id=data-and-tools}}.           
  
In another effort, an answer-based definition of RQE was proposed and tested \cite{BenAbacha_AMIA_2016}. The authors introduced a dataset of clinical questions and used a feature-based method that provided an Accuracy of 75\% on consumer health questions. We will call this dataset \textit{Clinical-QE}\footnote{\url{https://github.com/abachaa/RQE_Data_AMIA2016}}. Dos Santos et al. \cite{Santos:ACL2015} proposed a new approach to retrieve semantically equivalent questions combining a bag-of-words representation  with a  distributed vector representation created by a CNN and user data collected from  two  Stack  Exchange  communities. Lei et al. \cite{Lei:NAACL2016} proposed a  recurrent  and  convolutional  model  (gated  convolution)  to  map questions to their semantic representations. The models were pre-trained within an encoder-decoder framework.        
    
%%%========================================== 
\section{RQE Approaches and Experiments}
\label{sec3:RQE}  
%%%========================================== 

The choice of two methods for our empirical study is motivated by the best performance achieved by Logistic Regression in question-question similarity at SemEval 2017 (best system \cite{SimBow:SemEvalBest2017} and second best system \cite{SemEval2017-SecondT}), and the high performance achieved by neural networks on larger datasets such as SNLI \cite{snli:emnlp2015,Kim:2018,Chen:2018,Ghaeini:2018}. We first define the RQE task, then present the two approaches, and evaluate their performance on five different datasets.
 
%%==================== 
\subsection{Definition} 
\label{sec:RQEDef}     
%%%===================

In the context of QA, the goal of RQE is to retrieve answers to a new question by retrieving entailed questions with associated answers. We therefore define question entailment as: 
\begin{itemize}
\item a question $A$ \textbf{entails} a question $B$ if every answer to $B$ is also a \textbf{complete} or \textbf{partial} answer to $A$.   
\end{itemize} 
 
\noindent We present below two examples of consumer health questions $Ai$ and entailed questions $Bi$:  \\    
\textbf{Example 1} (each answer to the entailed question B1 is a \textit{complete} answer to A1): 
\begin{itemize}
\item A1: \textit{What is the latest news on tennitis, or ringing in the ear, I am 75 years old and have had ringing in the ear since my mid 5os. Thank you. }
\item B1: \textit{What is the latest research on Tinnitus?}   
\end{itemize} 
  
\textbf{Example 2} (each answer to the entailed question B2 is a \textit{partial} answer to A2): 
\begin{itemize}
\item A2: \textit{ My mother has been diagnosed with Alzheimer's, my father is not of the greatest health either and is the main caregiver for my mother. My question is where do we start with attempting to help our parents w/ the care giving and what sort of financial options are there out there for people on fixed incomes. }
\item B2: \textit{What resources are available for Alzheimer's caregivers? } 
\end{itemize} 
 
\noindent The inclusion of partial answers in the definition of question entailment also allows efficient relaxation of the contextual constraints of the original question $A$ to retrieve relevant answers from entailed, but less restricted, questions.                
     
% Example 2:
% \begin{itemize}
% \item A: Trisomy 13. I had a trisomy 13 mosaic infant and now my sister in law has just been diagnosed with a trisomy pregnancy. Is it possible both brothers may carry a translocation? 
% \item B: Is trisomy 13 inherited, and what is the pattern? 
% \end{itemize}

% Example 3:
% \begin{itemize}
% \item C: Is Tamsulosin gluten-free and where is it manufactured? Thank you. 
% \item D: Is Tamsulosin gluten-free? 
% \end{itemize}     
%%%========================================       
\subsection{Deep Learning Model}       
%%%========================================       
  
\begin{figure}[ht!]       
\centering              
  \includegraphics[scale=0.35]{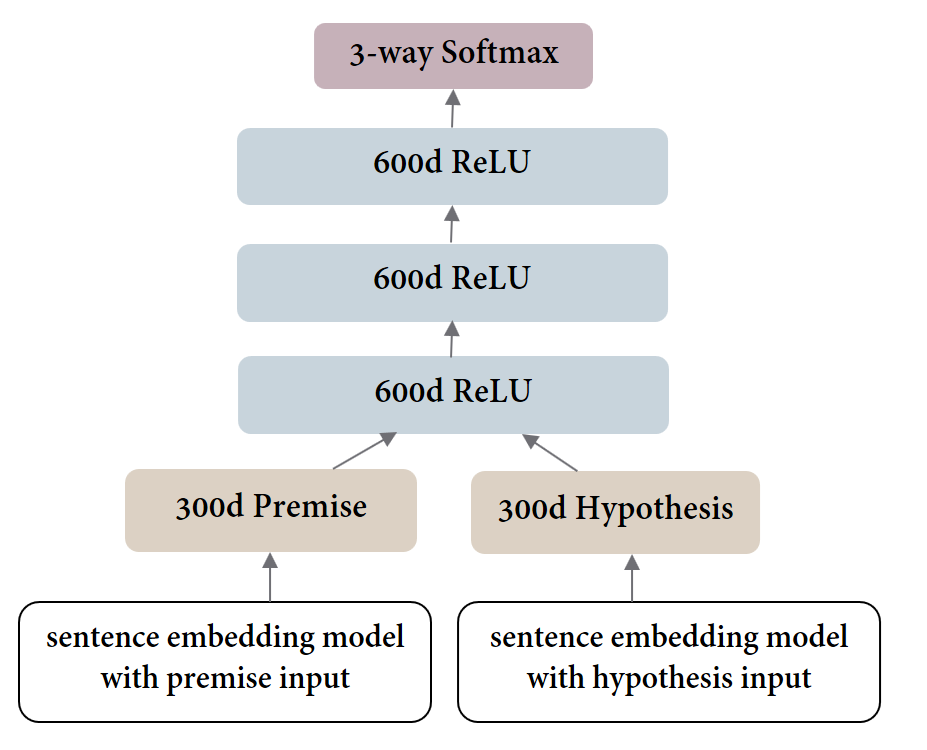}     
\caption{Neural Network Architecture}  
\label{fig:nn}     
\end{figure}

To recognize entailment between two questions $PQ$ (premise) and $HQ$ (hypothesis), we adapted the neural network proposed by Bowman et al. \cite{snli:emnlp2015}. Our DL model, presented in Figure~\ref{fig:nn}, consists of three 600d ReLU layers, with a bottom layer taking the concatenated sentence representations as input and a top layer feeding a softmax classifier. The sentence embedding model sums the Recurrent neural network (RNN) embeddings of its words. The word embeddings are first initialized with pretrained GloVe vectors. This adaptation provided the best performance in previous experiments with RQE data. 
     
GloVe\footnote{\url{https://nlp.stanford.edu/projects/glove}} is an unsupervised learning algorithm to generate vector representations for words \cite{GloveRef:2014}. Training is performed on aggregated word co-occurrence statistics from a large corpus, and the resulting representations show interesting linear substructures of the word vector space. We use the pretrained common crawl version with 840B tokens and 300d vectors, which are not updated during training. 
  
%%%========================================= 
\subsection{Logistic Regression Classifier} 
%%%=========================================   
 
In this feature-based approach, we use Logistic Regression to classify question pairs into entailment or no-entailment. Logistic Regression achieved good results on this specific task and outperformed other statistical learning algorithms such as SVM and Naive Bayes. In a preprocessing step, we remove stop words and perform word stemming using the Porter algorithm \cite{Porter:1980:Program} for all ($PQ$,$HQ$) pairs.     
  
We use a list of nine features, selected after several experiments on RTE datasets \cite{Book:Dagan:2013}. We compute five similarity measures between the pre-processed questions and use their values as features. We use Word Overlap, the Dice coefficient based on the number of common bigrams, Cosine, Levenshtein, and the Jaccard similarities. Our feature list also includes the maximum and average values obtained with these measures and the question length ratio (length($PQ$)/length($HQ$)). We compute a morphosyntactic feature indicating the number of common nouns and verbs between $PQ$ and $HQ$. TreeTagger \cite{Schmid1994} was used for POS tagging.  
  
For RQE, we add an additional feature specific to the question type. We use a dictionary lookup to map triggers to the question type (e.g. Treatment, Prognosis, Inheritance). Triggers are identified for each question type based on a manual annotation of a set of medical questions (cf. Section~\ref{sec:types}). This feature has three possible values: 2 (Perfect match between $PQ$ type(s) and $HQ$ type(s)), 1 (Overlap between $PQ$ type(s) and $HQ$ type(s)) and 0 (No common types).        
  
%%%==============================================
\subsection{Datasets Used for the RQE Study}        
%%%==============================================
\subsubsection{Training Datasets}     
We evaluate the RQE methods (i.e. deep learning model and logistic regression classifier) using two datasets of sentence pairs (SNLI and multiNLI), and three datasets of question pairs (Quora, Clinical-QE, and SemEval-cQA).      
  
The Stanford  Natural  Language  Inference corpus (SNLI) \cite{snli:emnlp2015} contains 569,037 sentence pairs written by humans based on image captioning. The training set of the MultiNLI corpus \cite{multi-nli:2017} consists of 393,000 pairs of sentences from five genres of written and spoken English (e.g. Travel, Government). Two other "matched" and "mismatched" sets are also available for development (20,000 pairs). Both SNLI and multiNLI consider three types of relationships between sentences: entailment, neutral and contradiction. We converted the contradiction and neutral labels to the same non-entailment class. 

The QUORA dataset of similar questions was recently published with 404,279 question pairs. We randomly selected three distinct subsets (80\%/10\%/10\%) for training (323,423 pairs), development (40,428 pairs) and test (40,428 pairs).   
 
The clinical-QE dataset \cite{BenAbacha_AMIA_2016} contains 8,588 question pairs and was constructed using 4,655 clinical questions asked by family doctors \cite{Ely_BMJ2000}. We randomly selected three distinct subsets (80\%/10\%/10\%) for training (6,870 pairs), development (859 pairs) and test (859 pairs).      

The question similarity dataset of SemEval 2016 Task 3B (SemEval-cQA) \cite{semEval_2016_ref} contains 3,869 question pairs and aims to re-rank a list of related questions according to their similarity to the original question. The same dataset was used for SemEval 2017 Task 3 \cite{semEval_2017_ref}. 

%%%================================================= 
\subsubsection{RQE Test Dataset}     
To construct our test dataset, we used a publicly shared set of Consumer Health Questions (CHQs) received by the U.S. National Library of Medicine (NLM), and annotated with named entities, question types, and focus \cite{Kilicoglu:LREC16,Kilicoglu:BMC18}. The CHQ dataset consists of 1,721 consumer information requests manually annotated with subquestions, each identified by a question type and a focus. 
 
First, we selected automatically harvested FAQs, from U.S. National Institutes of Health (NIH) websites, that share both the same focus and the same question type with the CHQs. As FAQs are most often very short, we first assume that the CHQ entails the FAQ. Two sets of pairs were constructed: (i) positive pairs of CHQs and FAQs sharing at least one common question type and the question focus, and (ii) negative pairs corresponding to a focus mismatch or type mismatch. For each category of negative examples, we randomly selected the same number of pairs for a balanced dataset. Then, we manually validated the constructed pairs and corrected the positive and negative labels when needed. The final RQE dataset contains 850 CHQ-FAQ pairs with 405 positive and 445 negative pairs. Table~\ref{tab:data} presents examples from the five training datasets (SNLI, MultiNLI, SemEval-cQA, Clinical-QE and Quora) and the new test dataset of medical CHQ-FAQ pairs.                   
   
\begin{table*}[ht]      
\centering           
\begin{tabular} {|p{2cm}|p{3cm}|p{1cm}|p{8cm}|} \hline   
\textbf{Datasets} & \textbf{Type/Domain} & \bf \# pairs & \bf Positive Examples (Entailment/Similarity)  \\\hline                                 
\textbf{SNLI} (2015) & Inference pairs of open-domain sentences.  & 550,152 (train) & \textbf{PS:} A child in a light and dark green ensemble sits in a chair in front of a typewriter looking off-camera. \textbf{HS:} A child sitting in front of a desk.     
 \\\hline       
\textbf{MultiNLI} (2017) & Inference pairs of open-domain sentences. & 392,702 (train)  & \textbf{PS:} On the island of the Giudecca, you'll find another of the great Palladio-designed churches (one of two in Venice), the Redentore. \textbf{HS:} There are two church in Venice that were designed by Palladio. \\\hline \hline           
\textbf{SemEval-cQA} (2016) & Similar questions from the Qatar Living forum. & 3,169 (train) & \textbf{PQ:} Books. Where can i donate books? \textbf{HQ:} english books. Where to buy english books? Is there a public library in doha? thanks \\\hline  
\textbf{Clinical-QE} (2016) & Entailment pairs of questions asked by doctors. & 8,588 & \textbf{PQ:} Patient is reluctant to take medications so I have been treating with smaller doses than I would with some other patients.  How do I control her hypertension and still get her cooperation? \textbf{HQ:} Patient reluctant to take medication. How to control hypertension and still get her cooperation?\\\hline       
\textbf{Quora} (2017) & Open-domain question similarity pairs. & 404,279 & \textbf{PQ:} I've been working out in the gym for the last three months but I'm not successful in gaining weight. Should I go for a mass gainer? Is it safe? \textbf{HQ:} I have been working out from few months but I am unable to gain mass/weight.Which mass gainer should I take? \\\hline  \hline    
\textbf{New Test Data (CHQs)} & Entailment pairs of consumer health questions. & 850 &   \textbf{PQ:} IHSS heart condition and WPW heart condition. Is there any way you could send me information on both these heart conditions? My son has to get tested for them eventually and I would just like information to understand the conditions of both of them more. \textbf{HQ:}  What is Wolff-Parkinson-White syndrome ?  \\\hline                                  
\end{tabular}            
\caption{Description of training and test datasets.}       
\label{tab:data}           
\end{table*}           
 
%%%=====================================================
\subsection{Results of RQE Approaches}  
\label{sec:rqeRes} 
%%%=====================================================

%======================= TABLE I ========================= 
\begin{table*}[ht]   
\centering    
\begin{tabular}  {|p{4.8cm}|p{2cm}|p{2.1cm}||p{2cm}|p{2.1cm}|}           
\hline           
\multirow{2}{*}{\textbf{Methods}}  &  \multicolumn{2}{ c|| }{\textbf{Textual Datasets}} & \multicolumn{2}{ |c| }{\textbf{Question Datasets}}  \\  \cline{2-5}           
& \textbf{SNLI (2015)} & \textbf{MultiNLI (2017) }  & \textbf{Quora (2017)} & \textbf{Clinical-QE (2016)} \\\hline      
\bf Neural Network (NN) & 79.50 & 73.71 & 81.34 & 71.45 \\\hline       
\bf NN + GloVe embeddings & \textbf{82.80} & \textbf{78.52} & \textbf{83.62}  & 93.12 \\\hline \hline      
\bf Logistic Regression + Features & 75.91 & 67.88 & 67.79 & \textbf{98.60} \\\hline      
\end{tabular} %           
\caption{Accuracy (\%) of Deep Learning and Machine Learning methods on four datasets SNLI, MultiNLI, Quora, and Clinical-QE.}  
\label{tab:resDL1}          
\end{table*}       
%======================= TABLE II =========================   
\begin{table*}[ht]     
\centering   
\begin{tabular}{|p{4.8cm}|p{2cm}|p{2cm}||p{2cm}|p{2cm}|}             
\hline    
\multirow{2}{*}{\textbf{Methods}}  &  \multicolumn{4}{ c| }{\textbf{Training Datasets}} \\  \cline{2-5}        
& \textbf{SNLI} & \textbf{MultiNLI}  & \textbf{Quora} & \cellcolor{gray!40} \textbf{Clinical-QE}   \\\hline           
\bf Neural Network (NN) & 48.94 & 54.59 & 52.35 & 48.71  \\\hline       
\bf NN + GloVe embeddings & 49.41 & 54.82 & 52.82 & 57.18  \\\hline \hline            
\cellcolor{gray!40} \textbf{Logistic Regression + Features} & 67.05 &  64.94 &  52.11 &  \textbf{73.18} \\\hline              
\end{tabular}      
\caption{Performance on 850 consumer health questions: Accuracy (\%) of Machine Learning and Deep Learning Methods.} 
\label{tab:resDL2}     
\end{table*}     
%====================== TABLE III =======================  
\begin{table*}[ht]  
\centering           
\begin{tabular}{|p{4.2cm}|p{1.9cm}||p{1.2cm}|p{0.8cm}|p{0.8cm}|p{0.8cm}|p{0.9cm}|p{0.9cm}|}        
\hline     
\bf Systems & \bf Test Sets & \bf Acc & \bf P & \bf R & \bf F1 & \bf MAP & \bf MRR \\\hline   
\textbf{Hybrid Method (Logistic Regression + Features + IR)} & cQA-16-Test &  \bf 80.57 & \bf  70.29 & \bf  72.10 & \bf  71.19 &  \bf  77.47 &  \bf  83.79 \\\hline  
\textit{cQA-B-2016 Best System \cite{semEval_2016_ref}}  & cQA-16-Test & \it 76.57 & \it 63.53 & \it 69.53 &  \it 66.39 & \it 76.70 & \it 83.02 \\\hline   \hline
\textit{cQA-B-2016 IR Baseline \cite{semEval_2016_ref}}  & cQA-16-Test & \it - & \it - & \it - &  \it - & \it 74.75 & \it 83.79 \\\hline   \hline 
\textbf{Hybrid Method (Logistic Regression + Features + IR)} & cQA-17-Test & \bf  67.27 & \bf  33.68 & 79.14 & \bf  47.25 & 44.66 & 48.08 \\\hline      
\textit{cQA-B-2017 Best System \cite{semEval_2017_ref}} & cQA-17-Test & \it  52.39 & \it  27.30 & \it  94.48 & \it  42.37 & \it  47.22 & \it  50.07 \\\hline    
\textit{cQA-B-2017 IR Baseline \cite{semEval_2017_ref}} & cQA-17-Test & \it - & \it - & \it  - & \it  - & \it 41.85 & \it  46.42 \\\hline   
\end{tabular}          
\caption{Results (\%) of the hybrid method (Logistic Regression + IR) on community QA datasets (cQA-Test 2016 and cQA-Test 2017). }      
\label{tab:res-16} 
\end{table*}      
%==============================================================
 
In the first experiment, we evaluated the DL and ML methods on SNLI, multi-NLI, Quora, and Clinical-QE. For the datasets that did not have a development and test sets, we randomly selected two sets, each amounting to 10\% of the data, for test and development, and used the remaining 80\% for training. For MultiNLI, we used the dev1-matched set for validation and the dev2-mismatched set for testing. 
 
Table~\ref{tab:resDL1} presents the results of the first experiment. The DL model with GloVe word embeddings achieved better results on three datasets, with 82.80\% Accuracy on SNLI, 78.52\% Accuracy on MultiNLI, and 83.62\% Accuracy on Quora. Logistic Regression achieved the best Accuracy of 98.60\% on Clinical-RQE. We also performed a 10-fold cross-validation on the full Clinical-QE data of 8,588 question pairs, which gave 98.61\% Accuracy.  

In the second experiment, we used these datasets for training only and compared their performance on our test set of 850 consumer health questions. Table~\ref{tab:resDL2} presents the results of this experiment. Logistic Regression trained on the clinical-RQE data outperformed DL models trained on all datasets, with 73.18\% Accuracy.        

To validate further the performance of the LR method, we evaluated it on question similarity detection. A typical approach to this task is to use an IR method to find similar question candidates, then a more sophisticated method to select and re-rank the similar questions. We followed a similar approach for this evaluation by combining the LR method with the IR baseline provided in the context of SemEval-cQA. The hybrid method combines the score provided by the Logistic Regression model and the reciprocal rank from the IR baseline using a weight-based combination:  
\begin{center}
$Hybrid score = LR\_score + w \times \frac{1}{IR\_rank}$  
\end{center}
\noindent The weight $w$ was set empirically through several tests on the cQA-2016 development set ($w=8.9$). Table~\ref{tab:res-16} presents the results on the cQA-2016 and cQA-2017 test datasets. The hybrid method (LR+IR) provided the best results on both datasets. On the 2016 test data, the LR+IR method outperformed the best system in all measures, with 80.57\% Accuracy and 77.47\% MAP (official system ranking measure in SemEval-cQA). On the cQA-2017 test data, the LR+IR method obtained 44.66\% MAP and outperformed the cQA-2017 best system in Accuracy with 67.27\%.        
 
%%%=====================================================
\subsection{Discussion of RQE Results}  
%%%=====================================================

When trained and tested on the same corpus, the DL model with GloVe embeddings gave the best results on three datasets (SNLI, MultiNLI and Quora). Logistic Regression gave the best Accuracy on the Clinical-RQE dataset with 98.60\%. When tested on our test set (850 medical CHQs-FAQs pairs), Logistic Regression trained on Clinical-QE gave the best performance with 73.18\% Accuracy.   
  
The SNLI and multi-NLI models did not perform well when tested on medical RQE data. We performed additional evaluations using the RTE-1, RTE-2 and RTE-3 open-domain datasets provided by the PASCAL challenge and the results were similar. We have also tested the SemEval-cQA-2016 model and had a similar drop in performance on RQE data.  
This could be explained by the different types of data leading to wrong internal conceptualizations of medical terms and questions in the deep neural layers.   
This performance drop could also be caused by the complexity of the test consumer health questions that are often composed of several subquestions, contain contextual information, and may contain misspellings and ungrammatical sentences, which makes them more difficult to process \cite{Dina_JAMIA_2016}. Another aspect is the semantics of the task as discussed in Section~\ref{sec:taskDef}. The definition of textual entailment in open-domain may not quite apply to question entailment due to the strict semantics. Also the general textual entailment definitions refer only to the premise and hypothesis, while the definition of RQE for question answering relies on the relationship between the sets of answers of the compared questions.         
  
%%%=====================================================
\section{Building a Medical QA Collection from Trusted Resources}   
\label{sec4:collection} 
%%%=====================================================  

A RQE-based QA system requires a collection of question-answer pairs to map new user questions to the existing questions with an RQE approach, rank the retrieved questions, and present their answers to the user.

\subsection{Method} 
  
To construct trusted medical question-answer pairs, we crawled websites from the National Institutes of Health\footnote{\url{www.nih.gov}} (cf. Section~\ref{sec:websites}). Each web page describes a specific topic (e.g. name of a disease or a drug), and often includes synonyms of the main topic that we extracted during the crawl.

We constructed hand-crafted patterns for each website to automatically generate the question-answer pairs based on the document structure and the section titles. We also annotated each question with the associated focus (topic of the web page) as well as the question type identified with the designed patterns (cf. Section~\ref{sec:types}).   
 
To provide additional information about the questions that could be used for diverse IR and NLP tasks, we automatically annotated the questions with the focus, its UMLS Concept Unique Identifier (CUI) and Semantic Type. We combined two methods to recognize named entities from the titles of the crawled articles and their associated UMLS CUIs: (i) exact string matching to the UMLS Metathesaurus\footnote{We used the umls-2017AA version. }, and (ii) MetaMap Lite\footnote{\url{https://metamap.nlm.nih.gov/MetaMapLite.shtml}} \cite{MetaMapLiteRef}. We then used the UMLS Semantic Network to retrieve the associated semantic types and groups.   

%=============================================
\subsection{Question Types} 
\label{sec:types}   
%=============================================

The question types were derived after the manual evaluation of 1,721 consumer health questions. Our taxonomy includes 16 types about Diseases, 20 types about Drugs and one type (Information) for the other named entities such as Procedures, Medical exams and Treatments. We describe below the considered question types and examples of associated question patterns.    
  
\begin{enumerate} 
\item  \textit{Question Types about Diseases (16)}: Information, Research (or Clinical Trial), Causes, Treatment, Prevention, Diagnosis (Exams and Tests), Prognosis, Complications, Symptoms, Inheritance, Susceptibility, Genetic changes, Frequency, Considerations, Contact a medical professional, Support Groups. \newline
Examples: 
\begin{itemize}
\item What research (or clinical trial) is being done for DISEASE? 
\item What is the outlook for DISEASE? 
\item How many people are affected by DISEASE? 
\item  When to contact a medical professional about DISEASE?
\item Who is at risk for DISEASE?
\item Where to find support for people with DISEASE? 
\end{itemize}

\item \textit{Question Types About Drugs (20):} Information, Interaction with medications, Interaction with food, Interaction with herbs and supplements, Important warning, Special instructions, Brand names, How does it work, How effective is it, Indication, Contraindication, Learn more, Side effects, Emergency or overdose, Severe reaction, Forget a dose, Dietary, Why get vaccinated, Storage and disposal, Usage, Dose. \newline
Examples: 
\begin{itemize}
\item Are there interactions between DRUG and herbs and supplements? 
\item What important warning or information should I know about DRUG?
\item Are there safety concerns or special precautions about DRUG?
\item What is the action of DRUG and how does it work?
\item Who should get DRUG and why is it prescribed?
\item What to do in case of a severe reaction to DRUG?  
\end{itemize} 
  
\item \textit{Question Type for other medical entities (e.g. Procedure, Exam, Treatment)}: Information.    
\begin{itemize}
\item What is Coronary Artery Bypass Surgery?
\item What are Liver Function Tests?  
\end{itemize}
\end{enumerate}

\begin{figure}[ht!]            
 \centering                
    \fbox{\includegraphics[scale=0.36]{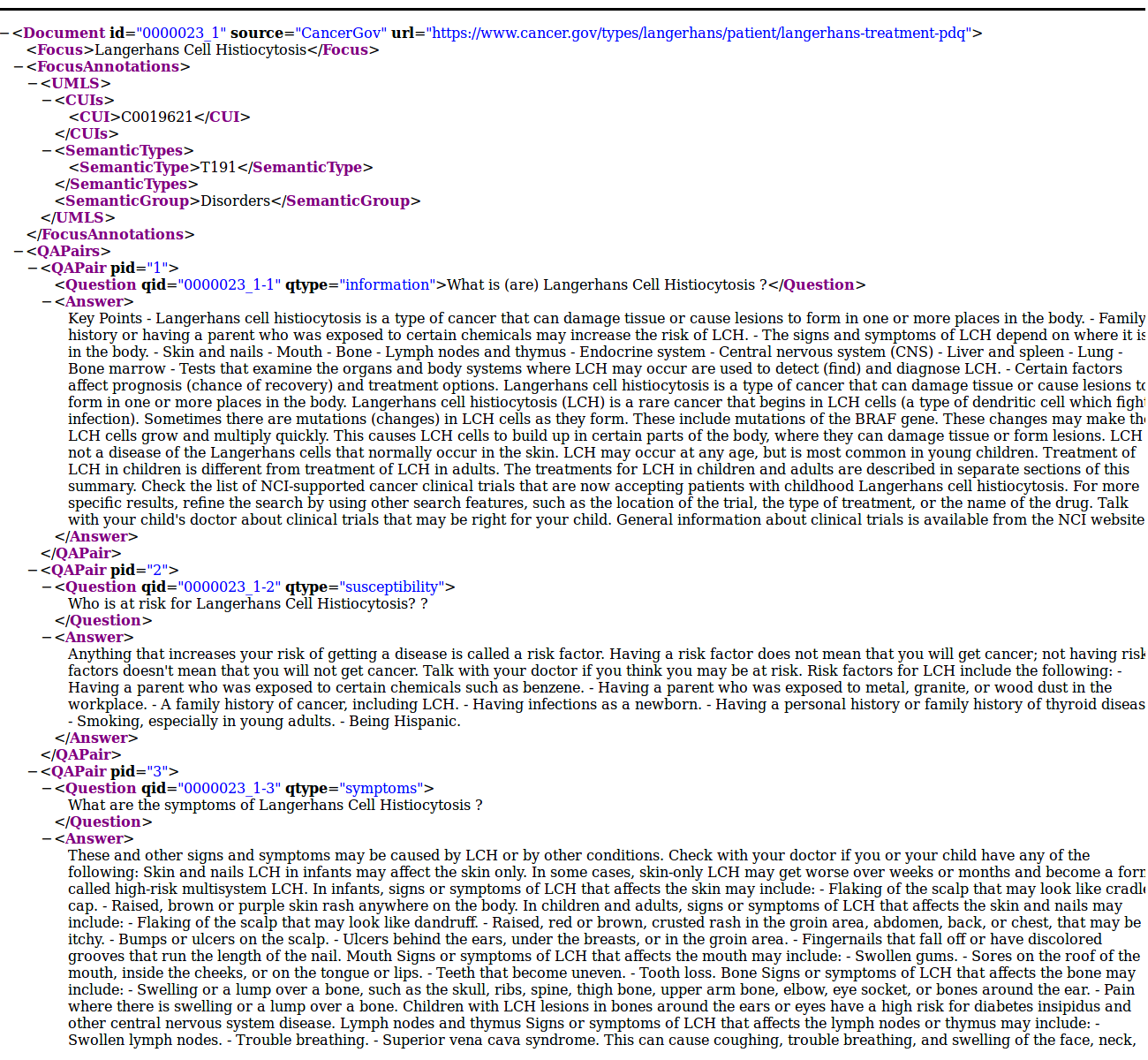}}         
 \caption{Examples of QA pairs generated from an article about \textit{Langerhans Cell Histiocytosis} (NCI). }    
 \label{fig:expQA2}      
\end{figure} 

\begin{figure}[ht!]           
\centering     
  \fbox{\includegraphics[scale=0.36]{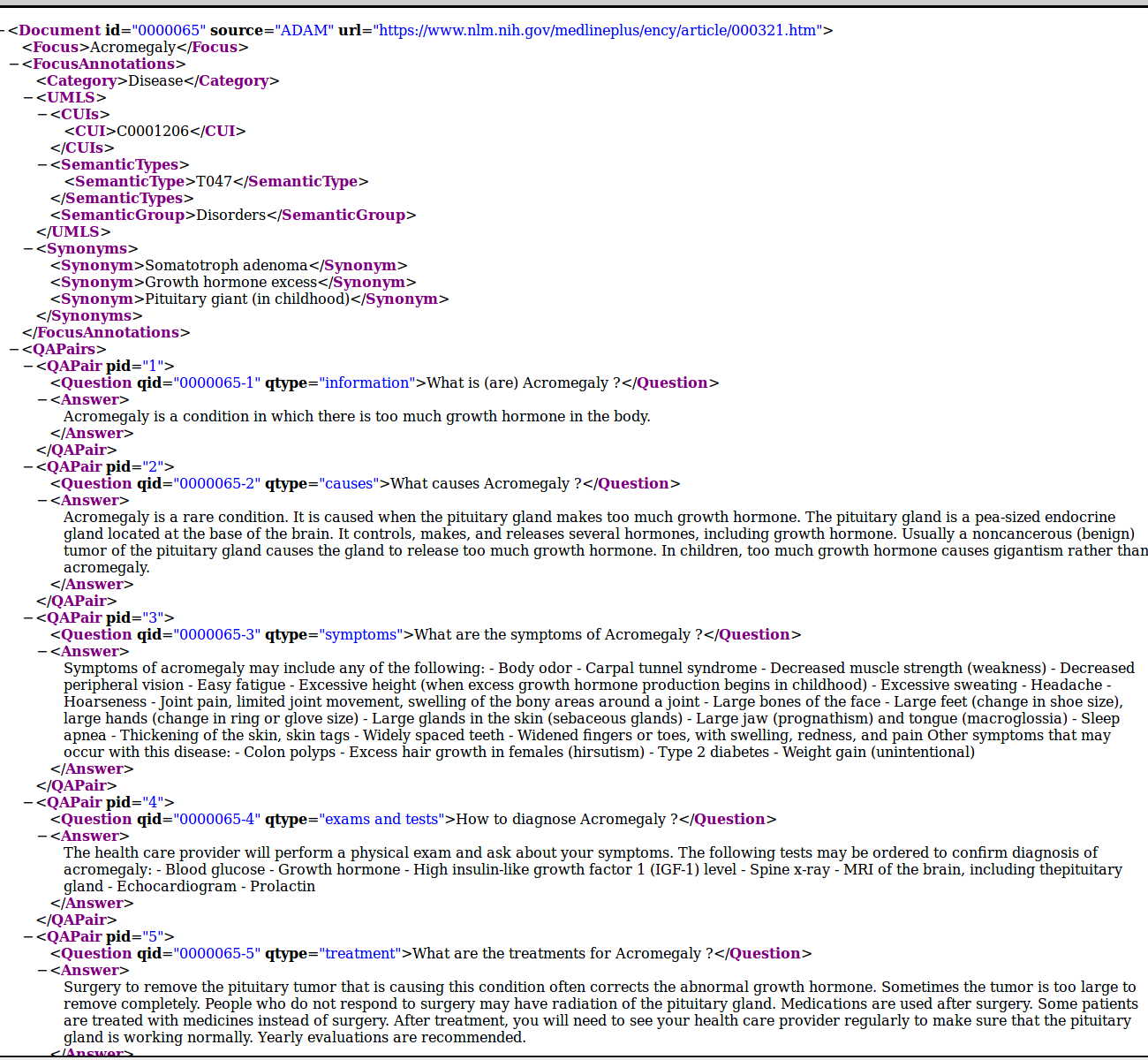}}  
\caption{Examples of QA pairs generated from an article about \textit{Acromegaly} (A.D.A.M encyclopedia). }      
\label{fig:expQA1}          
\end{figure}

\subsection{Medical Resources} 
\label{sec:websites} 
We used 12 trusted websites to construct a collection of question-answer pairs. For each website, we extracted the free text of each article as well as the synonyms of the article focus (topic). These resources and their brief descriptions are provided below:    
\begin{enumerate}       

\item National Cancer Institute (NCI) \footnote{\url{http://www.cancer.gov/types}}: We extracted free text from 116 articles on various cancer types (729 QA pairs). We manually restructured the content of the articles to generate complete answers (e.g. a full answer about the treatment of all stages of a specific type of cancer). Figure~\ref{fig:expQA2} presents examples of QA pairs generated from a NCI article.   %% 729!

\item Genetic and Rare Diseases Information Center (GARD)\footnote{\url{https://rarediseases.info.nih.gov/diseases}}: This resource contains information about various aspects of genetic/rare diseases. We extracted all disease question/answer pairs from 4,278 topics (5,394 QA pairs). %% 5,394 !  
 
\item Genetics Home Reference (GHR)\footnote{\url{https://ghr.nlm.nih.gov}}: This NLM resource contains consumer-oriented information about the effects of genetic variation on human health. We extracted 1,099 articles about diseases from this resource (5,430 QA pairs). %% 5,430 !  

\item MedlinePlus Health Topics\footnote{\url{https://medlineplus.gov/healthtopics.html}}: This portion of MedlinePlus contains information on symptoms, causes, treatment and prevention for diseases, health conditions and wellness issues. We extracted the free texts in summary sections of 981 articles (981 QA pairs). %% 981! 

\item National Institute of Diabetes and Digestive and Kidney Diseases (NIDDK) \footnote{\url{https://www.niddk.nih.gov/health-information}}: We extracted text from 174 health information pages on diseases studied by this institute (1,192 QA pairs). %% c t 1,233 ! 
  
\item National Institute of Neurological Disorders and Stroke (NINDS)\footnote{\url{https://www.ninds.nih.gov/Disorders/all-disorders}}: We extracted free text from 277 information pages on neurological and stroke-related diseases from this resource (1,104 QA pairs). %% 1,104

\item NIHSeniorHealth \footnote{\url{https://nihseniorhealth.gov/}}: This website contains health and wellness information for older adults. We extracted 71 articles from this resource (769 QA pairs). %% 769

\item National Heart, Lung, and Blood Institute (NHLBI) \footnote{\url{https://www.nhlbi.nih.gov/health/health-topics}}: We extracted text from 135 articles on diseases, tests, procedures, and other relevant topics on disorders of heart, lung, blood, and sleep (559 QA pairs). %% 559
 
\item Centers for Disease Control and Prevention (CDC) \footnote{\url{https://www.cdc.gov/diseasesconditions/}}: We extracted text from 152 articles on diseases and conditions (270 QA pairs).  %%  done! 

\item MedlinePlus A.D.A.M. Medical Encyclopedia\footnote{\url{https://medlineplus.gov/encyclopedia.html}}: This resource contains 4,366 articles about conditions, tests, and procedures. 17,348 QA pairs were extracted from this resource. Figure~\ref{fig:expQA1} presents examples of QA pairs generated from A.D.A.M encyclopedia.        %% ++ 

\item MedlinePlus Drugs\footnote{\url{https://medlineplus.gov/druginformation.html}}: We extracted free text from 1,316 articles about Drugs and generated 12,889 QA pairs.   %% ++ 

\item MedlinePlus Herbs and Supplements\footnote{\url{https://medlineplus.gov/druginfo/herb_All.html}}: We extracted free text from 99 articles and generated 792 QA pairs. %% ++ 
\end{enumerate}  
     
\noindent The final collection contains 47,457 annotated question-answer pairs about Diseases, Drugs and other named entities (e.g. Tests) extracted from these 12 trusted resources.

%===================================================
\section{The Proposed Entailment-based QA System}     
\label{sect:system}  
%=================================================== 
   
Our goal is to generate a ranked list of answers for a given Premise Question $PQ$ by ranking the recognized Hypothesis Questions $HQs$. Based on the RQE experiments above (Section~\ref{sec:rqeRes}), we selected Logistic Regression trained on the clinical-RQE dataset to recognize entailed questions and rank them with their classification scores.     
 
\subsection{RQE-based QA Approach}   
\label{sec:qaMethod}
 
\begin{figure*}[ht!]        
\centering           
\includegraphics[scale=0.4]{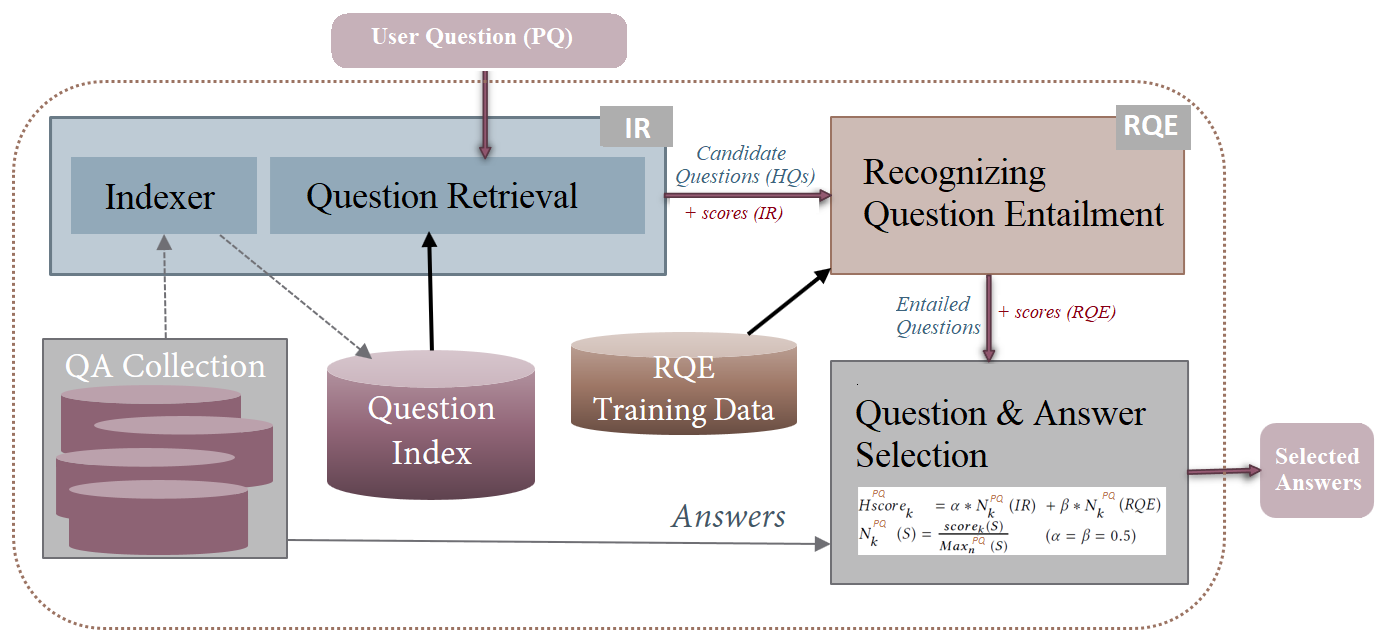}    
\caption{Overview of the RQE-based Question Answering System.}  
\label{fig:QAS}    
\end{figure*}

Classifying the full QA collection for each test question is not feasible for real-time applications. Therefore, we first filter the questions with an IR method to retrieve candidate questions, then classify them as entailed (or not) by the user/test question. Based on the positive results of the combination method tested on SemEval-cQA data (Section~\ref{sec:rqeRes}), we adopted a combination method to merge the results obtained by the search engine and the RQE scores. The answers are then combined from both methods and ranked using an aggregate score. Figure~\ref{fig:QAS}  presents the overall architecture of the proposed QA system. We describe each module in more details next.   

%%%=============================================== 

\subsection{Finding Similar Question Candidates} 

For each premise question $PQ$, we use the Terrier search engine\footnote{\url{http://terrier.org}} to retrieve $N$ relevant question candidates $\{HQ_j, j \in [1,N]\}$ and then apply the RQE classifier to predict the labels for the pairs ($PQ$,$HQ_j$).  

We indexed the questions of our QA collection without the associated answers. In order to improve the indexing and the performance of question retrieval, we also indexed the synonyms of the question focus and the triggers of the question type with each question. This choice allowed us to avoid the shortcomings of query expansion, including  incorrect or irrelevant synonyms and the increased execution time. The synonyms of the question focus (topic) were extracted automatically from the QA collection. The triggers of each question type were defined manually in the question types taxonomy. Below are two examples of indexed questions from our QA collection, with the automatically added focus synonyms and question type triggers:      
 
\begin{enumerate}
\item What are the treatments for Torticollis? 
\begin{itemize}
\item Focus: \textit{Torticollis}. Question type: \textit{Treatment}. 
\item Added focus synonyms: "Spasmodic torticollis, Wry neck, Loxia, Cervical dystonia". Added question type triggers: "relieve, manage, cure, remedy, therapy".  
\end{itemize}

\item What is the outlook for Legionnaire disease?
\begin{itemize}
\item Focus: \textit{Legionnaire disease}. Question Type: \textit{Prognosis}.  
\item Added focus synonyms: "Legionella pneumonia, Pontiac fever, Legionellosis". Added question type triggers: "prognosis, life expectancy".      
\end{itemize}     
\end{enumerate}    
 
\noindent The IR task consists of retrieving hypothesis questions $HQ_j$ relevant to the submitted question $PQ$. As fusion of IR result has shown good performance in different tracks in TREC, we merge the results of the TF-IDF weighting function and the In-expB2 DFR model \cite{Ref:Terrier}.      
   
Let $QL^V$ = {$HQ_1^V$, $HQ_2^V$, ..., $HQ_N^V$} be the set of $N$ questions retrieved by the first IR model $V$ and $QL^W$ = {$HQ_1^W$, $HQ_2^W$, ..., $HQ_N^W$} be the set of $N$ questions retrieved by the second IR model $W$. We merge both sets by summing the scores of each retrieved question $HQ_j$ in both $QL^V$ and $QL^W$ lists, then we re-rank the hypothesis questions $HQ_j$.             
    
%%%================================================   

\subsection{Combining IR and RQE Methods}   
  
The IR models and the RQE Logistic Regression model bring different perspectives to the search for relevant candidate questions. In particular, question entailment allows understanding the relations between the important terms, whereas the traditional IR methods identify the important terms, but will not notice if the relations are opposite. Moreover, some of the question types that the RQE classifier learns will not be deemed important terms by traditional IR and the most relevant questions will not be ranked at the top of the list. 

Therefore, in our approach, when a question is submitted to the system, candidate questions are fetched using the IR models, then the RQE classifier is applied to filter out the non-entailed questions and re-rank the remaining candidates.
   
Specifically, we denote $CL$ the list of question candidates $\{HQ_j, 1 \leq j \leq N\}$ returned by the IR system. The premise question $PQ$ is then used to construct N question pairs $\{ (PQ, HQ_j), 1 \leq j \leq N\}$. The RQE classifier is then applied to filter out the question pairs that are not entailed and re-rank the remaining pairs.  
 
More precisely, let $EL^{PQ}$ = $\{ HQ_1, HQ_2, \ldots, HQ_k\, \ldots \}$ in $CL$ be the list of selected candidate questions that have a positive entailment relation with a given premise question $PQ$. We rank $EL^{PQ}$ by computing a hybrid score $Hscore_{k}$ for each candidate question $HQ_k$ taking into account the score of the IR system $score_{k}(IR)$ and the score of the RQE system $score_{k}(RQE)$.      
 
\noindent For each system $S$ $\in{\{IR,RQE\}}$, we normalize the associated score by dividing it by the maximum score among the $N$ candidate questions retrieved by $S$ for $PQ$:            
\begin{itemize}
\item $Hscore_{k}^{PQ}=\alpha*Norm_k^{PQ}(IR) + \beta*Norm_k^{PQ}(RQE)$  
\item $Norm^{PQ}_k(S) =\frac{score_k(S)}{Max_N^{PQ}(S)}$ \hspace{0.7cm} $(\alpha=\beta=0.5)$      
\end{itemize}       
 
\noindent In our experiments, we fixed the value of $N$ to 100. This threshold value was selected as a safe value for this task for the following reasons:
\begin{itemize} 
\item Our collection of 47,457 question-answer pairs was collected from only 12 NIH institutes and is unlikely to contain more than 100 occurrences of the same focus-type pair.  
\item Each question was indexed with additional annotations for the question focus, its synonyms and the question type synonyms.
\end{itemize}

%===================================================
\section{Evaluating RQE for Medical Question Answering}    
\label{sec6:eval}
%===================================================
 
The objective of this evaluation is to study the effectiveness of RQE for Medical Question Answering, by comparing the answers retrieved by the hybrid entailment-based approach, the IR method and the other QA systems participating to the medical task at TREC 2017 LiveQA challenge (LiveQA-Med).        
  
\subsection{Evaluation Method}     

We developed an interface to perform the manual evaluation of the retrieved answers. Figure~5 presents the evaluation interface showing, for each test question, the top-10 answers of the evaluated QA method and the reference answer(s) used by LiveQA assessors to help judging the retrieved answers by the participating systems.     %%\ref{fig:evalInterface1} 

%=============================================== 
% \begin{figure*}[h]       
% \centering         
% \fbox{\includegraphics[width=15cm,height=11.2cm]{figures/Eval4.png}} 
% \caption{Interface used for the manual evaluation. For each test question, the reference answers used by LiveQA assessors were provided to help judging the retrieved answers by the QA systems.}     
% \label{fig:evalInterface}   
% \end{figure*}  
 
  \begin{figure}[h!]  
  \begin{adjustbox}{addcode={\begin{minipage}{\width}}{\caption{%
      Evaluation Interface: for each test question, the reference answers used by LiveQA assessors were provided to help judging the retrieved answers.  
      }\end{minipage}},rotate=90,center} 
      \includegraphics[scale=.42]{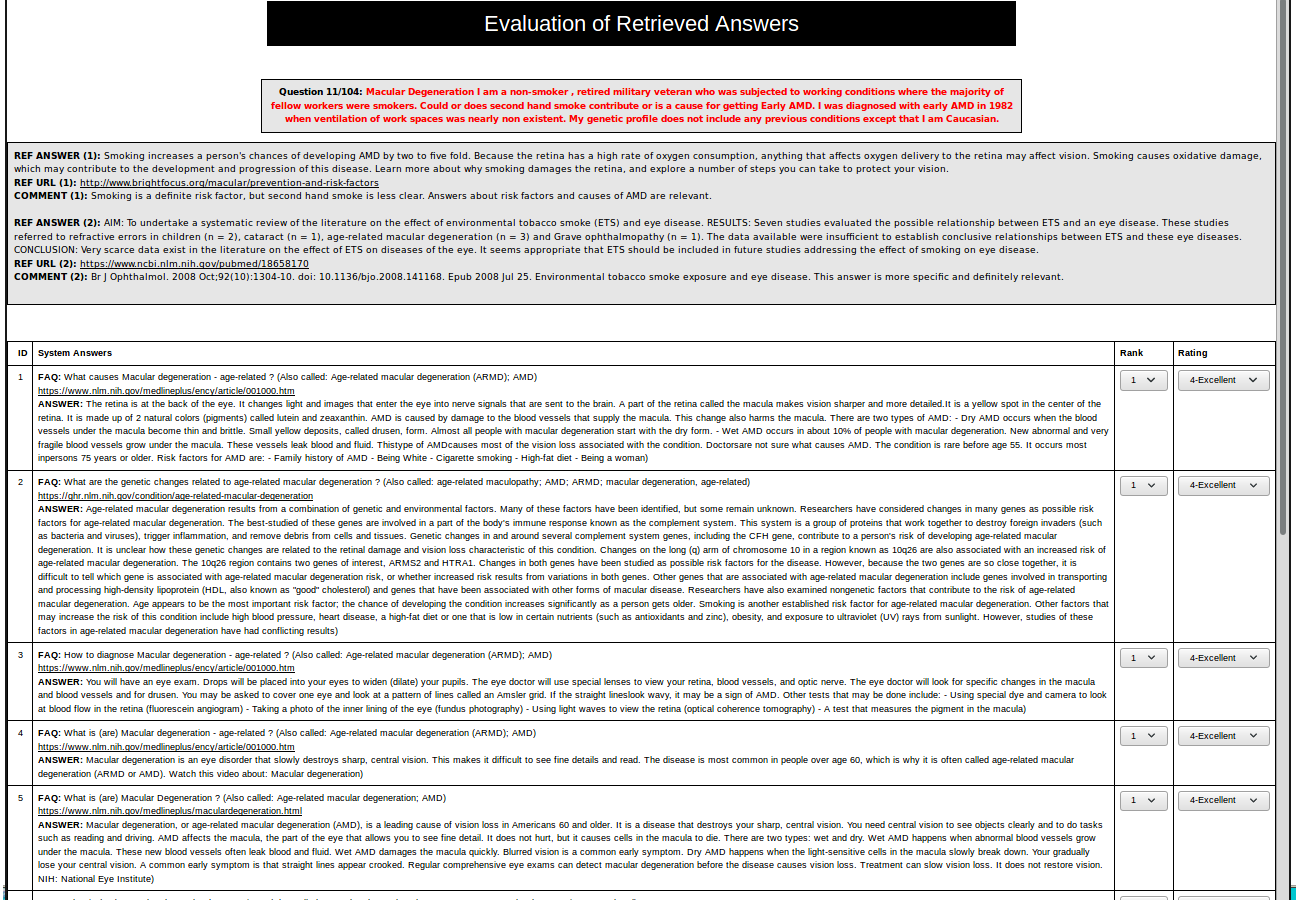}  
  \end{adjustbox}    
  \label{fig:evalInterface1}     
\end{figure}  
  
%===============================================
  
We used the test questions\footnote{\url{https://github.com/abachaa/LiveQA_MedicalTask_TREC2017}} of the medical task at TREC-2017 LiveQA \cite{liveQA-Med-overview-2017}.
These questions are randomly selected from the consumer health questions that the NLM receives daily from all over the world. The test questions cover different medical entities and have a wide list of question types such as Comparison, Diagnosis, Ingredient, Side effects and Tapering.   

\noindent For a relevant comparison, we used the same judgment scores as the LiveQA Track:  
\begin{itemize}  
\item Correct and Complete Answer (4) 
\item Correct but Incomplete (3) 
\item Incorrect but Related (2)
\item Incorrect (1)
\end{itemize}   
 
We evaluated the answers returned by the IR-based method and the hybrid QA method (IR+RQE) according to the same reference answers used in LiveQA-Med. The answers were anonymized (the method names were blinded) and presented to 3 assessors: a medical doctor (Assessor A), a medical librarian (B) and a researcher in medical informatics (C). None of the assessors participated in the development of the QA methods. Assessors B and C evaluated 1,000 answers retrieved by each of the methods (IR and IR+RQE). Assessor A evaluated 2,000 answers from both methods.         

Table~\ref{tab:IAA} presents the inter-annotator agreement (IAA) through F1 score computed by considering one of the assessors as reference. In the first evaluation, we computed the True Positives (TP) and False Positives (FP) over all ratings and the Precision and F1 score. As there are no negative labels (only true or false positives for each category), Recall is 100\%. We also computed a partial IAA by grouping the "Correct and Complete Answer" and "Correct but Incomplete" ratings (as Correct), and the "Incorrect but Related" and "Incorrect" ratings (as Incorrect).  The average agreement on distinguishing the Correct and Incorrect answers is 94.33\% F1 score. Therefore, we used the evaluations performed by assessor A for both methods. The official results of the TREC LiveQA track relied on one assessor per question as well.
 
\begin{table}[h]            
\centering                               
\begin{tabular}  {|p{1.6cm}|p{2cm}|p{2cm}|p{2cm}|p{2cm}|} \hline     
\multirow{2}{*}{\textbf{Assessors}}& \multicolumn{2}{ c }{\textbf{IAA}} & \multicolumn{2}{ |c| }{\textbf{Partial IAA}}   \\ \cline{2-5}           
& \bf P (\%) & \bf F1 (\%)  & \bf P (\%) & \bf F1 (\%)  \\\hline                             
 \textbf{A vs. B} & 80.80  & 89.38 & 90.13 & 94.81   \\\hline     
  \textbf{A vs. C} & 77.92  & 87.59 & 88.42 & 93.85  \\\hline        
\textbf{Average} & 79.36  & 88.48 & 89.27  & 94.33  \\\hline                           
\end{tabular}                      
\caption{Inter-Annotator Agreement (IAA) over all ratings in the manual evaluation of the retrieved answers. Partial IAA over two ratings "Correct" and "Incorrect". }                 
\label{tab:IAA}         
\end{table}         

\subsection{Evaluation of the first retrieved answer}  
  
We computed the measures used by TREC LiveQA challenges \cite{liveQA-overview-2015,liveQA-Med-overview-2017} to evaluate the first retrieved answer for each test question:      
\begin{itemize}   
\item avgScore(0-3): the average score over all questions, transferring 1-4 level
grades to 0-3 scores. This is the main score used to rank LiveQA runs.  
\item succ@i+: the number of questions with score i or above (i$\in${\{2..4\}}) divided by the total number of questions.  
\item prec@i+:  the number of questions with score i or above (i$\in${\{2..4\}}) divided by number of questions answered by the system.
\end{itemize}      
 
\begin{table*}[h]                 
\centering                                               
\begin{tabular}       {|p{2.2cm}|p{1.8cm}|p{1.9cm}||p{2cm}|p{2.3cm}|} \hline          
\textbf{Measures} & \bf IR-based System  & \textbf{IR+RQE System} & \bf LiveQA'17 Best Results & \bf LiveQA'17 Median Results \\\hline                        
\textbf{avgScore(0-3)} & 0.711  & \textbf{0.827} & 0.637  & 0.431 \\\hline   \hline       
\textbf{succ@2+} & 0.442 & \textbf{0.461}  & 0.392  & 0.245 \\\hline           
\textbf{succ@3+} & 0.192 &  0.25  & \textbf{0.265} & 0.142 \\\hline         
\textbf{succ@4+ } & 0.077 & \textbf{0.115} &  0.098 & 0.059 \\\hline \hline      
\textbf{prec@2+} & 0.46  & \textbf{0.475} & 0.404 & 0.331  \\\hline    
\textbf{prec@3+} & 0.2 & 0.257 & \textbf{0.273}  & 0.178 \\\hline          
\textbf{prec@4+ } & 0.08 &  \textbf{0.119} & 0.101 & 0.077  \\\hline       
\end{tabular}         
\caption{LiveQA Measures: Average Score (main score), Success@i+ and Precision@i+ on LiveQA'17 Test Data. Evaluation of the first retrieved answer for each question. \newline N.B. Evaluating the RQE System alone is not relevant as explained previously (Section~\ref{sec:qaMethod}).}    
\label{tab:resQA1}               
\end{table*}      
 
Table~\ref{tab:resQA1} presents the average scores, success and precision results. The hybrid IR+RQE QA system achieved better results than the IR-based system with 0.827 average score. It also achieved a higher score than the best results achieved in the medical challenge at LiveQA'17. Evaluating the RQE system alone is not relevant, as applying RQE on the full collection for each user question is not feasible for a real-time system because of the extended execution time. 
 
\subsection{Evaluation of the top ten answers}     

In this evaluation, we used Mean Average Precision (MAP) and Mean Reciprocal Rank (MRR) which are commonly used in QA to evaluate the top-10 answers for each question. We consider answers rated as ``Correct and Complete Answer'' or ``Correct but Incomplete'' as correct answers, as the test questions contain multiple subquestions while each answer in our QA collection can cover only one subquestion. 

MAP is the mean of the Average Precision (AvgP) scores over all questions. 

(1) $MAP = \frac{1}{Q}\sum\limits_{i=1}^Q AvgP_{i}$    
\begin{itemize}   
\item \textit{Q is the number of questions. $AvgP_{i}$ is the AvgP of the $i^{th}$ question.} 
\end{itemize} 

 $AvgP = \frac{1}{K}\sum\limits_{n=1}^K \frac{n}{rank_{n}}$    
\begin{itemize}    
\item \textit{K is the number of correct answers. $rank_{n}$ is the rank of $n^{th}$ correct answer.}
\end{itemize}     

MRR is the average of the reciprocal ranks for each question. The reciprocal rank of a question is the multiplicative inverse of the rank of the first correct answer. 

(2) $MRR = \frac{1}{Q}\sum\limits_{i=1}^Q \frac{1}{rank_{i}}$   

\begin{itemize}     
\item \textit{Q is the number of questions. $rank_{i}$ is the rank of the first correct answer for the $i^{th}$ question.}
\end{itemize}     
  
Table~\ref{tab:resQA2} presents the MAP@10 and MRR@10 of our QA methods. The IR+RQE system outperforms the IR-based QA system with 0.311 MAP@10 and 0.333 MRR@10.  
 
 \begin{table}[h]          
\centering                                  
\begin{tabular}{|p{5cm}|p{3cm}|p{3cm}|} \hline                
\textbf{Measures} & \textbf{IR-based System} & \textbf{IR+RQE System}  \\ \hline     
\textbf{Fully answered questions} & 29\% & 27\%   \\\hline 
\textbf{Correctly answered questions} & 51\% & 54\%  \\\hline \hline                 
\textbf{MAP@10} & 0.282 & 0.311  \\\hline                 
\textbf{MRR@10} & 0.281 & 0.333  \\\hline     
\end{tabular}            
\caption{Common Measures: MAP and MRR on LiveQA'17 Test Questions. Evaluation of top 10 answers for each question. }      
\label{tab:resQA2}      
\end{table}
   
%================================================================  
\subsection{Discussion of entailment-based QA for the medical domain}   
%================================================================ 

In our evaluation, we followed the same LiveQA guidelines with the highest possible rigor. In particular, we consulted with NIST assessors who provided us with the paraphrases of the test questions that they used to judge the answers. Our IAA on the answers rating was also high compared to related tasks, with an 88.5\% F1 agreement with the exact four categories and a 94.3\% agreement when reducing the categories to two: “Correct” and “Incorrect” answers.  Our results show that RQE improves the overall performance and exceeds the best results in the medical LiveQA'17 challenge by a factor of 29.8\%. This performance improvement is particularly interesting as: 
\begin{enumerate}[label=(\alph*)]
    \item Our answer source has only 47K question-answer pairs when LiveQA participating systems relied on much larger collections, including the World Wide Web.  
    \item Our system answered one subquestion at most when many LiveQA test questions had several subquestions. 
\end{enumerate} 
  
The latter observation, (b),  makes the hybrid IR+RQE approach even more promising as it gives it a large potential for the improvement of answer completeness. 

The former observation, (a), provides another interesting insight: restricting the answer source to only reliable collections can actually improve the QA performance without losing coverage (i.e., our QA approach provided at least one answer to each test question and obtained the best relevance score). 

In another observation, the assessors reported that many of the returned answers had a correct question type but a wrong focus, which indicates that including a focus recognition module to filter such wrong answers can improve further the QA performance in terms of precision. Another aspect that was reported is the repetition of the same (or similar) answer from different websites, which could be addressed by improving answer selection with inter-answer comparisons and removal of near-duplicates. Also, half of the LiveQA test questions are about Drugs, when only two of our resources are specialized in Drugs, among 12 sub-collections overall. Accordingly, the assessors noticed that the performance of the QA systems was better on questions about diseases than on questions about drugs, which suggests a need for extending our medical QA collection with more information about drugs and associated question types.     

We also looked closely at the private websites used by the LiveQA-Med annotators to provide some of the reference answers for the test questions. For instance, the ConsumerLab website was useful to answer a question about the ingredients of a Drug (COENZYME Q10). Similarly, the eHealthMe website was used to answer a test question asking about interactions between two drugs (Phentermine and Dicyclomine) when no information was found in DailyMed. eHealthMe provides healthcare big data analysis and private research and studies including self-reported adverse drug effects by patients. 

But the question remains on the extent to which such big data and other private websites could be used to automatically answer medical questions if information is otherwise unavailable. Unlike medical professionals, patients do not necessarily have the knowledge and tools to validate such information. An alternative approach could be to put limitations on medical QA systems in terms of the questions that can be answered (e.g. "What is my diagnosis for such symptoms") and build classifiers to detect such questions and warn the users about the dangers of looking for their answers online.    
  
More generally, medical QA systems should follow some strict guidelines regarding the goal and background knowledge and resources of each system in order to protect the consumers from misleading or harmful information.  Such guidelines could be based (i) on the source of the information such as health and medical information websites sponsored by the U.S. government, not-for-profit health or medical organizations, and medical university centers, or (ii) on conventions such as the code of conduct of the HON Foundation (HONcode) that addresses the reliability and usefulness of medical information on the Internet. 
 
Our experiments show that limiting the number of answer sources with such guidelines is not only feasible, but it could also enhance the performance of the QA system from an information retrieval perspective.   
  
%======================     
\section{Conclusion} 
\label{sec:conclusion}  
%======================     
  
In this paper, we carried out an empirical study of machine learning and deep learning methods for Recognizing Question Entailment in the medical domain using several datasets. We developed a RQE-based QA system to answer new medical questions using existing question-answer pairs. We built and shared a collection of 47K medical question-answer pairs\footnote{\url{https://github.com/abachaa/MedQuAD}}. Our QA approach outperformed the best results on TREC-2017 LiveQA medical test questions. The proposed approach can be applied and adapted to open-domain as well as specific-domain QA. Deep learning models achieved interesting results on open-domain and clinical datasets, but obtained a lower performance on consumer health questions. We will continue investigating other network architectures including transfer learning, as well as creation of a large collection of consumer health questions for training to improve the performance of DL models. Future work also includes exploring integration of a Question Focus Recognition module to enhance candidate question retrieval, and expanding our question-answer collection. 
    
\section{Acknowledgements}    

We thank Halil Kilicoglu (NLM/NIH) for his help with the crawling and the manual evaluation and Sonya E. Shooshan (NLM/NIH) for her help with the judgment of the retrieved answers. We also thank Ellen Voorhees (NIST) for her valuable support with the TREC LiveQA evaluation.     
  
\newpage
\noindent\textbf{Annex A: Case Study }\\ 
 
\noindent We consider the case of the question number 36 in the TREC-2017 LiveQA medical test dataset:  \\   
\textit{36. congenital diaphragmatic hernia. what are the causes of congenital diaphragmatic hernia? Can cousin marriage cause this? What kind of lung disease the baby might experience life long?}

\noindent This question was answered by 5 participating runs (vs. 8 runs for other questions), and all submitted answers were wrong (scores of 1 or 2). However, our IR-based QA system retrieved one excellent answer (score 4) and our hybrid IR+RQE system provided 3 excellent answers. \\ 
  
\noindent\textbf{A) TREC 2017 LiveQA-Med Participants' Results: }  
\begin{enumerate}   
\item 36 CMU-OAQA-CarnegieMellonUniversity 2  
\begin{itemize}
\footnotesize 
\item 36 2 Yes Opinion 2: CHERUBS is a non-profit organization created to help families affected by Congenital Diaphragmatic Hernia. If you visit the site you will learn all about CDH, read stories of patients (children and adults), see photos and more. Membership and information is free. http://www.cdhsupport.org My name is Dawn and I run CHERUBS. My son was born with CDH and I started CHERUBS 16 years ago. Now, we have over 3600 families in our group so I know there will be someone who truly understands what you're going through. If you'd like to talk or have questions, you can reach me at dawn.williamson@cdhsupport.org
\end{itemize}
  
\item 36 CMU-LiveMedQA-CarnegieMellonUniversity 1
\begin{itemize}
\footnotesize  
\item 36 1 Infants with congenital diaphragmatic hernias most commonly present with respiratory distress and cyanosis in the first minutes or hours of life, although a later presentation is possible. The respiratory distress can be severe and may be associated with circulatory insufficiency, requiring aggressive resuscitative measures. ; See Clinical Presentationfor more detail. Clinical Presentation ; As noted in Mortality/Morbidity, population-based studies show that congenital diaphragmatic hernia is diagnosed based on prenatal ultrasonography findings in approximately one half of affected infants. Infants may have a prenatal history ofpolyhydramnios. polyhydramnios Infants most commonly present with respiratory distress and cyanosis in the first minutes or hours of life, although a later presentation is possible. The respiratory distress can be severe and may be associated with circulatory insufficiency, requiring aggressive resuscitative measures. ; Infants with congenital diaphragmati
\end{itemize}

\item 36 ECNU\_ICA\_2-EastChinaNormalUniversity 1
\begin{itemize}
\footnotesize  
\item 36 1 Congenital Diaphragmatic Hernia (CDH) is a congenital abnormality that is often life-threatening because it can limit the growth of the lungs. The site of herniation is most commonly found on the left side in the hole originally designed for the entry of alimentary canal. Herniation is sometimes found near sternum. Contents * Defects It involves three major defects. * A failure of the diaphragm to completely close during development. * Herniation of the abdominal contents into the chest * Pulmonary Hypoplasia Early diagnosis and presentation This condition can often be diagnosed before birth and fetal intervention can sometimes help, depending on the severity of the condition.[1] Infants born with diaphragmatic hernia experience respiratory failure due to both pulmonary hypertension and pulmonary hypoplasia. The first condition is a restriction of blood flow through the lungs thought to be caused by defects in the lung. Pulmonary hypoplasia or decreased lung volume is directly related
\end{itemize}

\item 36 ECNU-EastChinaNormalUniversity 1 
\begin{itemize}
\footnotesize 
\item 36 1 The act of facial motion starts in the brain and travels through the facial nerves to the muscles in the face. These muscles then contract in response to a stimulus. Inside the skull, the facial nerve is a single nerve. Congenital facial palsy is a condition present at birth. Moebius syndrome is a congenital condition. In most cases the exact cause of congenital palsy is uncertain.
\end{itemize}

\item 36 ECNU\_ICA-EastChinaNormalUniversity 2 
\begin{itemize}
\footnotesize
\item 36 2 CHERUBS is a non-profit organization created to help families affected by Congenital Diaphragmatic Hernia. If you visit the site you will learn all about CDH, read stories of patients (children and adults), see photos and more. Membership and information is free. http://www.cdhsupport.org My name is Dawn and I run CHERUBS. My son was born with CDH and I started CHERUBS 16 years ago. Now, we have over 3600 families in our group so I know there will be someone who truly understands what you're going through. If you'd like to talk or have questions, you can reach me at dawn.williamson@cdhsupport.org
\end{itemize} 
\end{enumerate}    

\noindent\textbf{B) Our IR-based QA System}:    
%36_Answer1	4-Excellent
\begin{itemize}   
\footnotesize  
\item Answer S1-1 (MedlinePlus\footnote{\url{https://www.nlm.nih.gov/medlineplus/ency/article/001135.htm}}: What causes Diaphragmatic hernia ? (Also called: Hernia - diaphragmatic; Congenital hernia of the diaphragm (CDH)) ANSWER: A diaphragmatic hernia is a rare defect. It occurs while the baby is developing in the womb. Because the diaphragm is not fully developed, organs, such as the stomach, small intestine, spleen, part of the liver, and the kidney, may take up part of the chest cavity.  CDH most often involves only one side of the diaphragm. It is more common on the left side. Often, the lung tissue and blood vessels in the area do not develop normally either. It is not clear if the diaphragmatic hernia causes the underdeveloped lung tissue and blood vessels, or the other way around.  40\% of babies with this condition have other problems as well. Having a parent or sibling with the condition increases the risk.
\end{itemize} 

\noindent\textbf{C) Our IR+RQE QA System}:      
% 36_Answer5	4-Excellent
% 36_Answer7	4-Excellent
% 36_Answer8	4-Excellent
\begin{enumerate}
\footnotesize  
\item Answer S2-1 (from GHR\footnote{\url{https://ghr.nlm.nih.gov/condition/congenital-diaphragmatic-hernia}}): What are the genetic changes related to congenital diaphragmatic hernia ? (Also called: congenital diaphragmatic defect) ANSWER: Congenital diaphragmatic hernia has many different causes. In 10 to 15 percent of affected individuals, the condition appears as a feature of a disorder that affects many body systems, called a syndrome. Donnai-Barrow syndrome, Fryns syndrome, and Pallister-Killian mosaic syndrome are among several syndromes in which congenital diaphragmatic hernia may occur. Some of these syndromes are caused by changes in single genes, and others are caused by chromosomal abnormalities that affect several genes.  About 25 percent of individuals with congenital diaphragmatic hernia that is not associated with a known syndrome also have abnormalities of one or more major body systems. Affected body systems can include the heart, brain, skeleton, intestines, genitals, kidneys, or eyes. In these individuals, the multiple abnormalities likely result from a common underlying disruption in development that affects more than one area of the body, but the specific mechanism responsible for this disruption is not clear.  Approximately 50 to 60 percent of congenital diaphragmatic hernia cases are isolated, which means that affected individuals have no other major malformations.  More than 80 percent of individuals with congenital diaphragmatic hernia have no known genetic syndrome or chromosomal abnormality. In these cases, the cause of the condition is unknown. Researchers are studying changes in several genes involved in the development of the diaphragm as possible causes of congenital diaphragmatic hernia. Some of these genes are transcription factors, which provide instructions for making proteins that help control the activity of particular genes (gene expression). Others provide instructions for making proteins involved in cell structure or the movement (migration) of cells in the embryo. Environmental factors that influence development before birth may also increase the risk of congenital diaphragmatic hernia, but these environmental factors have not been identified.
%%%%%%%%%%%%%%%%%%%%%%%%%%%%%%%%%%%%%%%%%%%%%%%%%%
\item Answer S2-2 (from GHR): Is congenital diaphragmatic hernia inherited ? (Also called: congenital diaphragmatic defect): Isolated congenital diaphragmatic hernia is rarely inherited. In almost all cases, there is only one affected individual in a family.  When congenital diaphragmatic hernia occurs as a feature of a genetic syndrome or chromosomal abnormality, it may cluster in families according to the inheritance pattern for that condition.
%%%%%%%%%%%%%%%%%%%%%%%%%%%%%%%%%%%%%%%%%%%%%%%%%%   
\item Answer S2-3 (from MedlinePlus\footnote{\url{https://www.nlm.nih.gov/medlineplus/ency/article/001135.htm}}) What is the outlook for Diaphragmatic hernia ? (Also called: Hernia - diaphragmatic; Congenital hernia of the diaphragm (CDH)): The outcome of surgery depends on how well the baby's lungs have developed. It also depends on whether there are any other congenital problems. Most often the outlook is good for infants who have a sufficient amount of working lung tissue and have no other problems.  Medical advances have made it possible for over half of infants with this condition to survive. The babies survived will often have ongoing challenges with breathing, feeding, and growth.
\end{enumerate}

%%% EVAL 2
%\begin{itemize}
%\item 36	36_Answer1	1	4-Excellent
% \item 36	36_Answer2	1	3-Incomplete
% \item 36	36_Answer3	1	3-Incomplete
% \item 36	36_Answer4	1	3-Incomplete
% \item 36	36_Answer5	1	2-Related
% \item 36	36_Answer6	1	2-Related
% \item 36	36_Answer7	1	2-Related
% \item 36	36_Answer8	1	2-Related
% \item 36	36_Answer9	1	2-Related
% \item 36	36_Answer10	1	3-Incomplete
%\end{itemize}

%%% EVAL 3
%\begin{itemize}
% \item 360 36_Answer1	1	3-Incomplete
% \item 36	36_Answer2	1	3-Incomplete
% \item 36	36_Answer3	1	3-Incomplete
% \item 36	36_Answer4	1	2-Related
%\item 36	36_Answer5	1	4-Excellent
% \item 36	36_Answer6	1	2-Related
%\item 36	36_Answer7	1	4-Excellent
%\item 36	36_Answer8	1	4-Excellent
% \item 36	36_Answer9	1	2-Related
% \item 36	36_Answer10	1	2-Related
%\end{itemize} 

\newpage     
\textbf{Annex B:} List of test questions used in the medical task at TREC LiveQA-Med\footnote{\url{https://github.com/abachaa/LiveQA_MedicalTask_TREC2017}}  
    
\begin{enumerate}  
\footnotesize 
\item What are the references with noonan syndrome and polycystic renal disease	
\item Gluten information. Re:NDC\# 0115-0672-50  Zolmitriptan tabkets 5mg.  I have celiac disease \&amp; need to know if these contain gluten,  Thank you!  
\item amphetamine salts 20 mg. are they gluten free 
\item vdrl positive. vdrl positive patients please tell me what are he doing . Diagnosis and precaution. 
\item how much glucagon. How much glucose is in my GlucaGen HypoKit ?  Just curious, I know that there is enough because I have used it.  Thank you very much
\item ANESTHESIA EFFECT ON FXTAS PERSONS. Does Anesthesia given during a operation severely hurt, or damage a brain for FXTAS patient?  The operation would be for hip replacement! Thank you very much
\item DVT. Can a birth control called Ocella cause DVT?  My daughter experiences pains cramping,redness and swelling in her thigh and also really bad huge blood clots during her menstrual cycles after she was prescribed Osella for birth control. Also these syntoms worsened after she gave birth. This has been happening for a year now should she see discuss this with her doctor right away?
\item medication question. I have had a bad UTI for 3 months I have taken cipro 7 times uti returns days after I oomplete I need a new prescription but the doctors here can figure out what to give me as I am allergic to penicillin and allergic to dairy products wich is a filler in many drugs.  Please please give me some idea of what I can get my dr; to prescribe
\item can a streptococcus infection cause an invasive disease like wegeners or the symptoms of wegeners? 
\item Diabetes and pain control. How can I narrow my search to find information regarding pain(joint) medication suitable to use with a person who has diabetes type 2.
\item Macular Degeneration. I am a non-smoker , retired military veteran who was subjected to working conditions where the majority of fellow workers were smokers. Could or does second hand smoke contribute or is a cause for getting  Early AMD. I was diagnosed with early AMD in 1982 when ventilation of work spaces was nearly non existent. My genetic profile does not include any previous conditions except that I am Caucasian.
\item molar pregnancy. is conception a requirement of a molar pregnancy. if so, when ?
\item symptoms and diagnosis. My son is being tested now to see if he has hnpp and after reading about the disease, it occurred to me that all my trouble with my hands could have been this and not arthritis. I have had both hands operated on several times, with some success, but continue with swelling in my hands and feet/ankles and soreness and stiffness. Would it be easy to think a patient has arthritis?
\item Yes my wife has been dianosed with giant cell vasculitis Our doctors are not clear about this so im asking for help 	From you . She has vomited something like coffee grounds and swelling in her feet and legs is really bad.migranes and face swelling to.no blood clots but nothing to go on so please help if u can thank u [NAME] [CONTACT]
\item can't find an answer. I was diagnosed with Fibromyalgia with chronic pain along with some other things and my blood work showed that I was missing a chromosone. How would I find out if I have a genetic for of Fibromyalgia?
\item cant use site. I want to find a doctor who specializes in burning mouth syndrome and that could be in many specialities, I cannot understand how to do this on your website. 		
\item estradiol 75g patch. Can I stop using the patch only been on it 4.5 months
\item Ear Wax. I sometimes drop Peroxide into the ear and let it bubble for a couple of minutes, then use warm water to flush it out. is there harm?		
\item Sevoflurane. 	I work in a hospital, and a question recently came up regarding the stability of Sevoflurane once it has been opened. Does Sevoflurane expire within a particular timeframe or is the product still effective until the expiration date listed on the bottle?
\item ODD. Would like to learn more about condition on ODD
\item Beckwith-Wieddeman Syndrome. I would like to request further knowledge on this specific disorder.
\item CITROBACTOR FREUNDII. Does ciprofaxin work well? Is there a better drug if so what.		
\item Hi I have a toddler 22 months and he was long exposure to car seat when he was infant and developed a flat head by then that was resolved, but since then he seems like his back is not well, he only sleep on his tummy, he hates to lay down on his back , he has a bad sitting position when on his car seat and other thing, I was wondering if he may need an evaluation to avoid further damage to his back, please let me know what kinf of doctor should I see, cause his pedi. Dr. Does not has any concerns about it. Thanks.
\item Ear Ache. My son was treated for otitis media on [DATE]. Pains started the previous night.  He is taking amoxicillin and antipyrine-benzocaine 5.5\%-1.5\% ear drops. This morning he woke up with a bit of blood drainage. Is that normal?
\item reaction to these 2 drugs. I would like to know if there are any drug reaction between Carvedilol 25 mg to Hydrslazine50 mg
\item mixing medications. Hello, I would like to know if taking Dicyclomine 20mg, phentermine can have a adverse effect? 
\item Dementia. Is dementia genetically passed down or could anyone get it
\item Is there a "sleep apnea surgery". I've heard that there is , but have never found a doctor that does this. My husband has been on C-pap for two years but has not been able to keep it on for more then 2 hours. He is not overweight, has had a stroke at 40 years old and double by-pass at 50  years old. Otherwise he follows doctors orders and has no other problems.  Thank you for your time, [NAME]
\item Diahrrea. I take Loperamide for chronic diahrrea. Then I stop it for about 2 days so I can have a bowel movement. But then the stool is really soft and there were a few times I almost didn't make to the bathroom. Is there a way for a happy medium
\item about uveitis. IS THE UVEITIS, AN AUTOIMMUNE DISEASE?
\item Customer Service Request. How much urine does an average human bladder hold - in ounces?
\item Amlodipine. I am taking Amlodipine and it has caused my pause rate to be very high. Is there a weaning process when you stop taking Amlodipine and start atenolol?   I am taking 5 mg of amlodipine and will be taking 50 mg of atenolol?
\item Shingles vacine. At what age should you get the Shingles shot. My children are in theur late 30's, early 40's, all three had bad cases of chicken pox as children.
\item very worry and need advise . dear sir     i had car accident 2 months a go . other person blood splash one me and i saw a lot of blood on my hand and some on face . not sure about eye . i didn't wash it immediatly and until 15 minute later then i washed it .     am i risk hiv ?       thank you .
\item Swan NDC 0869-0871-43. I found 4 cases of expired (04/2010) Hydrogen Peroxide.  How do I safely dispose of this product?
\item congenital diaphragmatic hernia. what are the causes of congenital diaphragmatic hernia? Can cousin marriage cause this? What kind of lung disease the baby might experience life long?
\item shingles. need to know about the work place and someone having shingles, especially while handling food.
\item ClinicalTrials.gov - Question - general information. My question to you is: what is the reason that there is very little attentions is to Antiphosoholipid Syndrome? To find the causes and possibly some type of cure for us who struggle with this auto-immune blood disorder? I guess that since it is female directed (9-1 female to male) that no one important enough has died from APS? Oh, by the way, I'm a 58 year old man!
\item side efectes to methadone. i jest started taking methadone and have confusion my face itches
\item methylprednisolole. unable to fine info on the above med
\item Simvastatin. Why is it recommended that this medicine be taken in the evening? Any harm in taking it in the morning?
\item Prednisone. My husband has been on Prednisone for almost a year for a Cancer treatment he had. He started at 30mg and stayed on 10mg until a couple weeks ago. The prednisone was causing other side effects. He reduced down to 5mg for a couple days and now has been off the prednisone for a week. How long should we expect this drug to stay in his system. He is really experiencing chills/fever/abdominal pain..are these common when coming off this drug? Is there anything else we should expect?
\item Does electrical high voltage shock cause swallowing problems in the near future ??
\item calcitonin salmon nasal spray. I picked up a bottle of above but noted it had NOT been refrigerated for at least the 3 days since Rx was filled. Box and literature state "refrigerate until opened." Pharmacist insisted it was ok "for 30 days" although I said that meant after opening. Cost is \$54.08 plus need to know if it will be as effective as should be. Thank you.
\item Schmorl's Nodes. I am trying to obtain information on subject matter.  
\item Topic not covered. What exactly is sleep paralysis?
\item Article on Exercise for Impaired - Overweight - Asthmatics. I just found the site through the article on breathing difficulty. My frustration is, WHAT exercises to do with asthma?  Today I walked out of the door of the house to take a walk, a beautiful, cool, blustery, sunny day.  Suddenly I couldn't catch my breath, my upper chest felt 'heavy', and I had to go back inside and sit down for a while.  I'm no exercise weenie, before a couple of bad accidents (car crashes waiting at stop lights!) I used to play softball, volleyball, basketball, even a bit of rugby, I was a dancer and a weight-exerciser, a bicyclist, rollerblader, tree climber. Even today, BP is typically 110-120 over 70-80, heart rate is fine too.  Is this just asthma?  WHAT exercises can I do, safely? Sure, we ALL get it, exercise is good for us.  Just which ones?  LOTS of us with asthma would love help here.  Thanks.
\item hi. I can't find this I take Ambien every night for sleep. I want to no how long before I go to bed am I supposed to take it.
\item bundle blockage. could you please tell me what a bundle blockage is. what are the symptoms. what is usually done for this?    Thank you
\item I have an infection in gums...dentist prescribed Cephalexin 500mg...Is this ok to take even though I am ALLERGIC TO PENICILLAN? 
\item Arrhthmia. can arrhythmia occurs after ablation? What is the success rate of Ablation? During my Holter test it was found that my Heart rate fluctuates from 254 to 21. How do you rate the situation?
\item fildena. Hello I was wondering if fildena is truly like Viagra. I'm trying to find an alternative since my insurance no longer will cover Viagra for what ever reason. Would like to know all relavent information regarding fildena. About all I've found is that it is not fda approved so any information would be helpful thanks
\item Nph. I am interested in a movement class.  I have nph and can find no help with exercise or support group.  Any ideas from [LOCATION] Med?
\item ClinicalTrials.gov - Question - specific study. Can Low dose naltrexone be used for treatment of severe depression? 
\item How do you catch hepatitis?
\item Jock Itch. I have Jock itch, and I have read through your symptoms. I wanted know if small lumps under the skin around the scrotum area is a symptoms as well? Should I be concerned?
\item What are the causes of rib cage pain? And and the remedy
\item The hantavirus can lead to death?
\item Is there always elevated temperature associated with appendicitis?
\item Frequency. My urologist has prescribed Oxybutinin, 5 mg tablets, NOT in the ER version. I understood they were to be taken once a day, but he has prescribed twice. Is this the correct recommended dosage, or should the prescription have been for once a day?
\item I have a question for your website an it seems to have difficulty answering . I  want to know if you take Gabamentine an hydrocodene together what would happen? ; if I take them separately it don't work.
\item Drug interactions. is it safe to take diclofenac when taking lisinopril or aleve or extra-strength Tylenol?
\item patau sydrome/ trisomy 13. i was wondering the condition of trisomy progresses over time (gets worse as they become older) also, how to diognose the disorder thank you!
\item quinine in seltzer water. Is it ok to drink quinine in seltzer water to ease leg cramps? If so, what would be the correct "dosage"?  It has a nasty taste but it does ease leg cramps.  Thank you.
\item Mite Infestation. Please inform me of the recommended treatment and prevention protocol for mite infestation in humans, particularly one that is non-toxic or has minimal side effects.
\item meds taken with wine at dinnertime. Is it safe to take my meds with wine at dinnertime?
\item neo oxy. pkease send me the indication and usage info for this powder. NEO-OXY 100/100 MR - neomycin sulfate and oxytetracycline hydrochloride powder
\item What are the reasons for Hypoglycemia in newborns.. and what steps should a pregnent take to avoid this.
\item diverticulitis. can diverticulosis or diverticulitis be detected by a cat scan if there is no infection at that time?
\item CAUSE OF A COLD. i UNDERSTAND CONTAGION AND TRANSFERRENCE OF COLD \&quot;GERMS\&quot; WHY ARE SOME PEOPLE AFFECTED AND OTHERS NOT?
\item Janumet XR 50mg/1000mg- 1 daily. Doctor prescribed for type 2 diabetes w/Metformin 500 mg 2 times daily. Pharmacy refused to fill stating overdose of Metformin. Who is right \&amp; what is maximum daily dosage of Metformin? Pharmacy is a non-public pharmacy for a major city employer plan provided for employees only.
\item SSPE. My son is 33years of age and did not have the measles vaccination.Could SSPE occur at this age or in the future?
\item ClinicalTrials.gov - Question - general information. My granddaughter was born with Klippel-Tranaunay Syndrome...There is very little information about this.  We are looking for the current research and treatments available.  She is 5 months old now and her leg seems to be most affected.  We want to get her help as soon as possible to address the symptoms and treat her condition.
\item Iron Overdose. Um...i took 25 iron pills...what do i do...this was last night
\item Inherited Ricketts. Mother has inherited ricketts. Passing A child but not B child. How likely would B child pass it on their child?
\item Medicare Part B coverage. I suffer with acute fibromyalgia (sp?) and the various drugs my doctor has prescribed for me have little if any effect in helping to control the pain. My doctor has since given me a prescription to have massage therapy which she thought medicare would cover. However, when checking with medicare, it turns out that it does not! Can you suggest any other type of treatment?
\item Homozygout MTHFR A1298C Health Issues and long term prognosis? What is your position on Homozygout MTHFR A1298C Health Issues and long term prognosis?
\item Vitamin D intake. Can high doses of Vitamin D (50,000 IUs per week) cause flatulence, among other possible effects?  And is such a high dose safe to raise very low levels of Vitamin D in the body?
\item Shingles. I am looking for information on how to prevent a shingles outbreak.
\item my father age 65 his always leg pain which use medicine
\item CVID. I have recently been diagnosed with CVID. As a person with thyroid a thyroid tumor greater than 1 cm. and several thyroid cysys I am concerned about cancer. What are the current stats. The tumor is being monitored by my endocrinologist.
\item diabete. whats diabete
\item wellbutrin xl 150. how to taper off
\item Periodic liver tests for patients on Lipitor. I was told at one point that anyone on Lipitor should have blood screening for liver damage every 6 months. Is this currently still the recommendation? NOTE: Although I am in recovery, I also have a history of alcoholism.
\item Hi  I have heard in order to get benefit of calcium, it should take with Magnesium, is that right ? I bought Calcium Castco ( Kirkland ) brand with D please let me know if is good for me because I am osteoporosis . Please help me . thanks
\item Testing for EDS. I would like to know if you can point me in the direction of a laboratory in Southern California, Specifically San Bernardino County or LA County or even Riverside County that does genetic testing for EDS or Osteogenesis Imperfecta and do you know if the two diseases are similiar in symptoms? Thank you for you help and time.
\item Consultation. Hello! I have acute chronic cervicitis caused by tampon use. It took a year and a half of treatment (medicines, cauterization), but the symptoms do not stop inflammation and analyzes not determined that bacteria produce inflammation me. I wonder if, despite not being a sexual cervicitis infection, it can cause infertility. And that's what makes a tampon that causes inflammation. Thank you very much!
\item trisomy 7. i am a 32 y/o who has had 4 miscarriages in the past 19 months. Upon my last DNC two weeks ago revealed a genetics study diagnosis of the baby having trisomy 7. could you offer me any information on this? could this have been maternal or paternal? is this something i would be a carrier of? What are the causes? i have tried but haven't found much information
\item metformin. Does metformin cause high blood pressure?
\item aclidinium. is this a steroid?  is there a problem using this if there is a possibility of the need for cataract surgery within the next 12 months?
\item intestines  digestion and sbsorption. kindly explain the general effects of smoking or rather the effects of nicotine to digestion and arbsoption
\item sswollen feet ankles legs I have fibromyalgia. When suffering from fibromyalgia will that cause swollen in your body . The swollen started yesterday
\item Can cancer spread through blood contact. Sir, after giving an insulin injection to my uncle who is a cancer patient the needle accidentally pined my finger. Is there a problem for me? Plz reply.
\item Plantar Fasiciitis. Is it true that more likely than not that Plantar fasiciitis could be aggravated by a consistancy of weight bearing activities?    Are there other forms of aggravation? if so will you please inform me.
\item CAN LIPNODES AND OR LIVER CANCER BE DETECTED IN A UPPER GI. CAN LIPNODES AND OR LIVER CANCER BE DETECTED IN A UPPER GI
\item abscess teeth. Can an abscess teeth cause a heart attack
\item ischemic syncope stroke diagonses. define?
\item SPECIFY COMPONENTS. COENZYME Q10(100-mg). WHAT ARE THE COMPONENTS OF THIS MEDICINE? IS IT USEABLE FOR MUSLIMS?
\item Autoimmune illness. What doctor specializes in testing for and treatment of autoimmune illness?
\item how does effextor cause ED and what is the mimimum amount that causes ED. I take effexor.  Is there a mimimum amount that will not cause ED
\item NSAIDS as a potential cause of ED. How long has this non prescription drug been implicated in erectile  dysfunction?
\item i want to know more about aeortic stenosis
\item What can cause white cells ti uprate
\item Glimepiride Storage \&amp; Alloweable Excursion Data. Can you please provide Glimepiride storage \&amp; allowable temperature excursion data, specifically for pharmacy and warehouse storage
\end{enumerate}        

\newpage 
\bibliographystyle{abbrvnat}    
\bibliography{RQE2019} 
   
\end{document}